%% file: iclr2027_conference.tex
\documentclass{article} 
\usepackage{iclr2027_conference,times}

\input{math_commands.tex}

\usepackage{microtype}
\usepackage{hyperref}
\usepackage{url}
\usepackage[misc]{ifsym} 
\usepackage{booktabs, subcaption, siunitx}
\usepackage{xcolor}
\usepackage{wrapfig}
\usepackage{graphicx, tabularx}
\usepackage{multirow}
\usepackage{array}
\usepackage{dsfont}
\usepackage{capt-of}
\usepackage{adjustbox}
\usepackage[table]{xcolor}
\usepackage[most]{tcolorbox}
\newcommand{\best}[1]{\textbf{#1}}
\newcommand{\second}[1]{\underline{#1}}

\usepackage{cuted}
\usepackage{capt-of}
\usepackage{array}

\newcommand{\logo}{\raisebox{-4pt}{\includegraphics[width=1.8em]{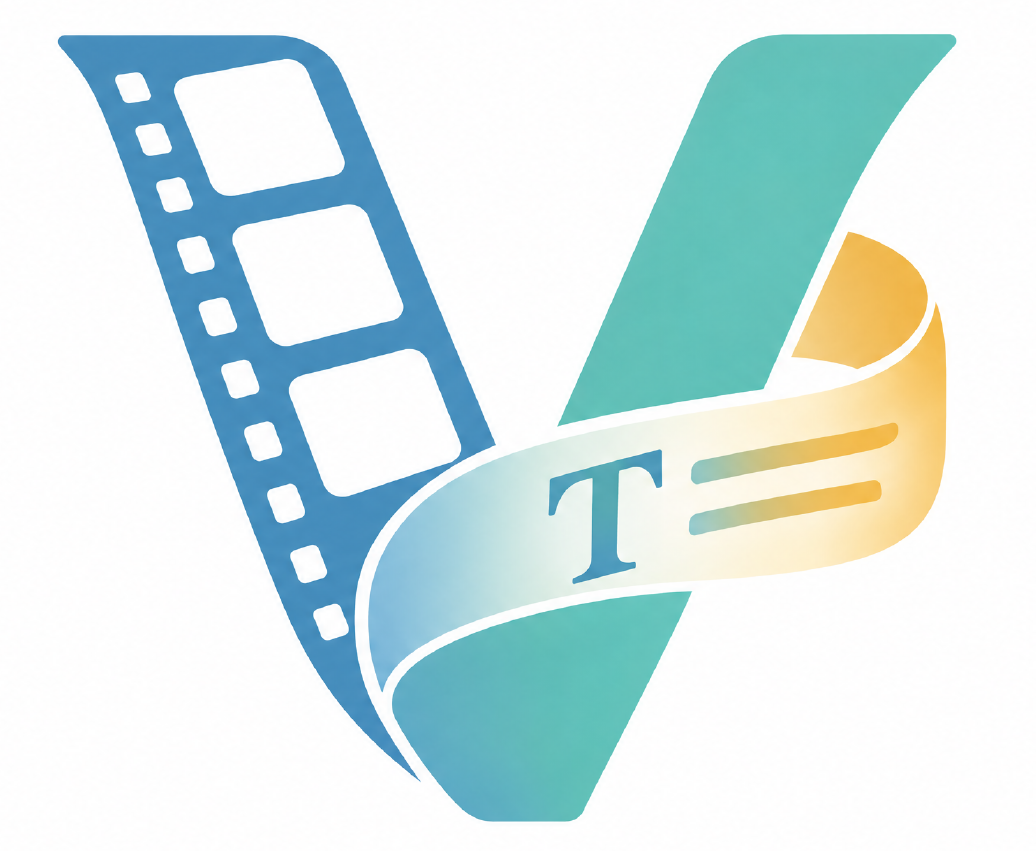}}}

\title{\logo~Beyond Legibility: Benchmarking Visual Text Rendering and In-Place Editing in Unified Video Generation}

\author{
Ziying Zhang\textsuperscript{\rm 1}\thanks{These authors contributed equally.},
Litao Li\textsuperscript{\rm 1}\footnotemark[1],
Junchao Liao\textsuperscript{\rm 1}\footnotemark[1],
Tianyi Zeng\textsuperscript{\rm 2},
Siyu Zhu\textsuperscript{\rm 3},
Long Qin\textsuperscript{\rm 1},
Zhenghao Zhang\textsuperscript{\rm 1}\thanks{Corresponding Author}
\\
\textsuperscript{\rm 1}Alibaba Group \quad
\textsuperscript{\rm 2}Shanghai Jiao Tong University \quad
\textsuperscript{\rm 3}Fudan University \\
\texttt{jiala.zzy@alibaba-inc.com} \quad
\textsuperscript{\dag}\texttt{zhangzhenghao.zzh@alibaba-inc.com}
}

\iclrfinalcopy 
\begin{document}

\maketitle

\begin{center}
\includegraphics[width=\textwidth]{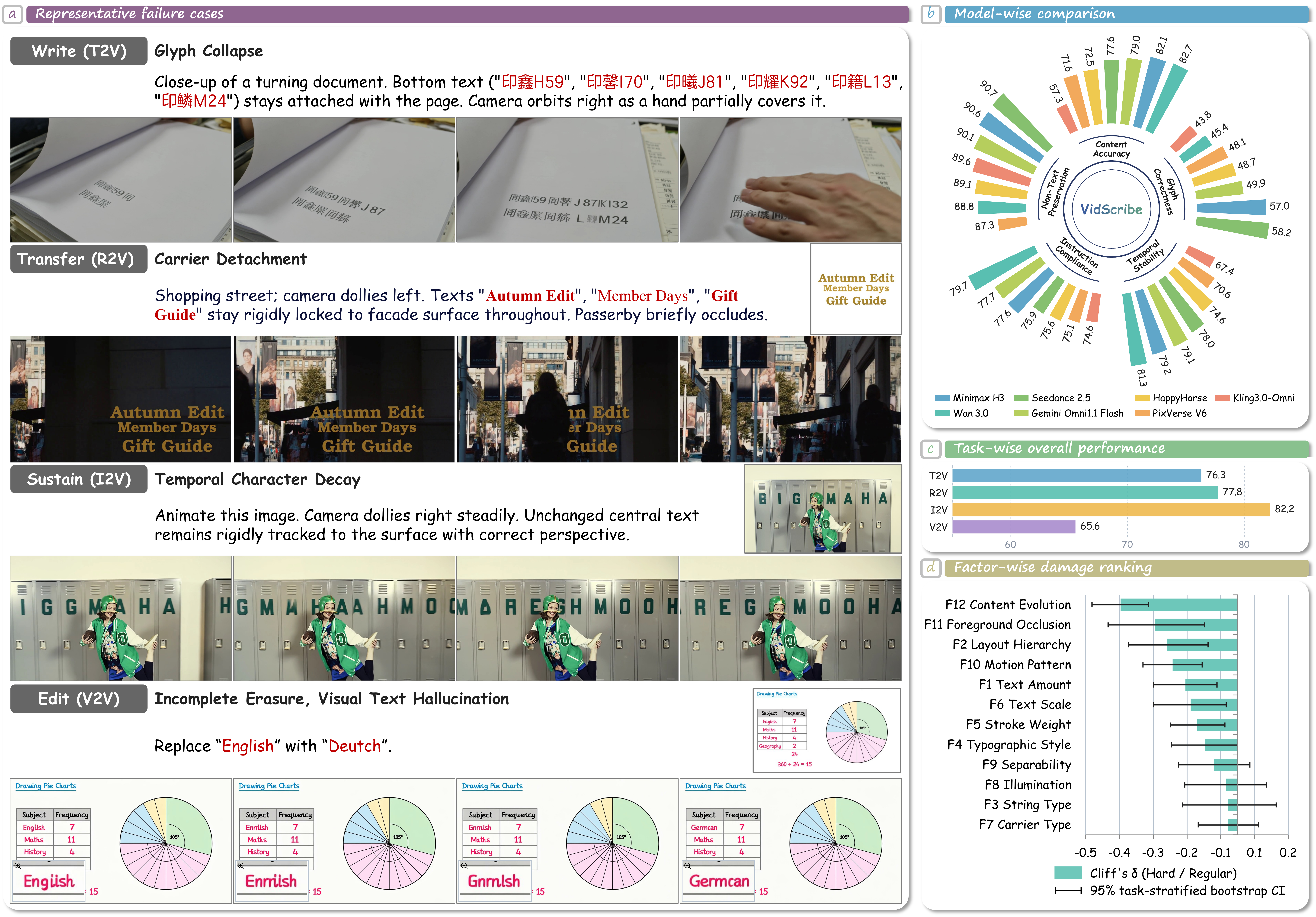}
\captionof{figure}{Motivating failure cases and headline findings of VidScribe. 
(a) Representative failure cases across T2V, R2V, I2V, and V2V. 
(b) Model-wise comparison across shared metrics, showing decoupling between Content Accuracy and Glyph Correctness. 
(c) Task-wise overall performance aggregated over all 7 commercial unified video generators, highlighting strong task asymmetry. 
(d) Factor-wise damage ranking aggregated over all 7 commercial unified video generators.}
\label{fig:teaser}
\end{center}

\begin{abstract}
A video can exhibit convincing motion and photorealism yet still fail immediately when visual text collapses. Unlike generic scene content, visual text is exceptionally unforgiving in video generation, where minor stroke corruption, temporal instability, or editing errors instantly break legibility and realism. Existing benchmarks overlook this challenge by treating text as incidental scene content or relying on static OCR metrics that ignore temporal dynamics. We introduce \textbf{VidScribe}, a unified diagnostic benchmark for visual text across 4 core generation regimes: writing from language (T2V), transferring text identity from reference (R2V), sustaining consistency under dynamics (I2V), and editing localized text within video (V2V). VidScribe provides 803 human-verified samples instantiated over a 12-axis conditionally orthogonal factor space spanning \textit{Intrinsic Text Properties}, \textit{Physical Imaging Conditions}, and \textit{Temporal Behavior}. To enable reliable evaluation, we build a track-grounded, gated suite with 11 shared metrics and 2 task-specific probes that enforce strict measurability conditions. Extensive benchmarking across 11 representative commercial and open-source systems reveals that video text capability is non-monolithic, with a clear decoupling between content recognition and stroke-level glyph correctness. Performance is highly task-asymmetric, with I2V sustaining text most reliably and V2V editing emerging as the primary bottleneck. Counter-intuitively, performance degradation is more concentrated on a small subset of text-centric structural and temporal factors than on adverse imaging conditions. Further diagnostic probes show that visual reference improves glyph and typographic fidelity rather than content accuracy, while localized editing struggles to isolate target text without corrupting undeclared source text. Beyond evaluation, VidScribe also provides an actionable training signal, where benchmark-aligned preference optimization measurably improves visual text generation. Data is available at https://huggingface.co/datasets/Vicky0720/VidScribe.
\end{abstract}

\section{Introduction}

Recent advances in video generation have substantially improved visual quality, motion realism, and prompt following~\citep{seedance202625, wan2026wan27, happyhorse2026, kuaishou2024kling}. Yet visual text rendering and manipulation remain poorly understood and inadequately evaluated. Unlike general scene content, text has almost zero tolerance for generative error: distortions that are acceptable on natural objects immediately break character legibility and realism. Missing strokes, unstable glyph shapes, temporal flicker, or residual artifacts after editing make a video look flawed at a glance. Visual text is therefore not a cosmetic detail, but a strict test of whether a foundation model can sustain fine-grained structural precision and temporal consistency.

Current evaluation protocols fail to assess this capability. Mainstream video benchmarks~\citep{han2025video, zheng2025vbench, sun2025t2v, wei2026msavbench} focus on overall visual quality or broad prompt alignment, largely treating text as incidental scene content. Conversely, existing visual text benchmarks~\citep{tuo2024anytext, zeng2024textctrl, liu2026scenevtg++, shu2025visual} remain predominantly image-based, focusing on static OCR accuracy or single-image editing. As a result, they miss failure modes unique to dynamic video, such as temporal character decay, carrier detachment, and incomplete erasure during editing. While existing benchmarks assess whether a video looks plausible overall, they cannot determine whether a model can reliably render, sustain, and edit text over time.

This diagnostic gap becomes clear when visual text is examined across 4 practical generation settings. Text-to-video~(T2V) requires \textbf{writing} legible text from language prompts alone. Reference-to-video~(R2V) entails \textbf{transferring} text identity from an exemplar image into a novel scene. Image-to-video~(I2V) focuses on \textbf{sustaining} pre-existing text under camera motion and scene dynamics. Finally, Video-to-video~(V2V) demands \textbf{editing} designated text while keeping the surrounding video intact. These settings probe distinct yet complementary capabilities, and as shown in Figure~\ref{fig:teaser}(a), they exhibit markedly different failure patterns.

To address this problem, we introduce \textbf{VidScribe}, a unified diagnostic benchmark for visual text across T2V, R2V, I2V, and V2V. VidScribe provides \textbf{803} human-verified samples built on a 12-axis factor space spanning \textit{Intrinsic Text Properties}, \textit{Physical Imaging Conditions}, and \textit{Temporal Behavior}. These axes are conditionally orthogonal, and an automated feasibility filter removes physically impossible combinations to enable controlled factor isolation. To evaluate model outputs reliably, we construct a track-grounded, gated suite of 11 shared metrics and 2 task-specific probes. The shared metrics evaluate \textit{Text Fidelity}, \textit{Temporal Stability}, \textit{Instruction Compliance} on typographic attributes, and \textit{Non-Text Preservation}, while the \textit{Task-Specific Probes} directly measure localized editing residue and source text preservation that general metrics miss.

Benchmarking 11 commercial and open-source video generators~\citep{kuaishou2024kling, wan2026wan27, seedance202625, minimax2026h3, geminiomni, PixVerse, happyhorse2026, wu2025hunyuanvideo, xiao2026joyai,lin2026kiwi,hacohen2026ltx} reveals three consistent findings. First, video text capability is non-monolithic. As shown in Figure~\ref{fig:teaser}(b), models exhibit a clear decoupling between lexical content recognition and stroke-level glyph correctness, where correct spelling frequently coexists with broken glyph topology. Second, performance is highly task-asymmetric: models sustain text most reliably in I2V, but struggle severely with V2V editing, while reference-guided transfer (R2V) consistently outperforms text-only writing (T2V), as shown in Figure~\ref{fig:teaser}(c). Third, performance degradation is more concentrated on a small subset of text-centric structural and temporal factors than on adverse imaging conditions. The largest drops arise from content evolution, foreground occlusion, and layout hierarchy, as illustrated in Figure~\ref{fig:teaser}(d).

Further targeted diagnostic probes clarify the technical reasons behind these behaviors. Under matched T2V--R2V comparisons, providing a visual exemplar substantially improves glyph structure and typographic style realization, yet yields virtually no gain in character-level spelling accuracy. Meanwhile, V2V editing exposes clear limits in localized control, where models struggle to isolate and modify target text regions without leaving residual artifacts or corrupting undeclared source text. These results suggest that current models handle text preservation and appearance transfer more reliably than localized text rewriting under spatio-temporal constraints.

Finally, VidScribe provides an actionable training signal. By converting our metric diagnostics into verifiable preference pairs, we perform preference optimization on an open-source video generator~\citep{wan2025wan}, achieving measurable gains in visual text fidelity and temporal preservation.

Our contributions are summarized as follows:
\begin{itemize}
    \item We present \textbf{VidScribe}, the first unified benchmark for visual text in video across T2V, R2V, I2V, and V2V, covering the 4 core settings of write, transfer, sustain, and edit.
    \item We establish a 12-axis conditionally orthogonal factor space paired with a track-grounded, gated evaluation suite of 11 shared metrics and 2 task-specific probes to evaluate text under diverse physical-imaging and temporal conditions.
    \item We conduct a systematic empirical study across 11 prominent video generation systems, showing that video text capability is non-monolithic, task-asymmetric, and acutely sensitive to text-centric structural and temporal dynamics rather than imaging hazards.
    \item We demonstrate the downstream value of VidScribe for preference-based alignment, showing that its diagnostic signals can be leveraged to improve visual text generation.
\end{itemize}

\section{Related Work}

\subsection{Visual Text Generation and Editing}
Visual text generation and editing have evolved from synthetic compositing and GAN-based pipelines to diffusion-based synthesis. Early approaches relied on synthetic rendering engines or GANs for text compositing, style transfer, and text removal~\citep{gupta2016synthetic, long2020unrealtext, krishnan2023textstylebrush, liu2022don}, but frequently suffered from blurry glyphs and boundary artifacts. Recent diffusion models have substantially improved 2D text generation by incorporating explicit structural priors, such as glyph masks, character bounding boxes, and OCR representations~\citep{chen2023textdiffuser, yang2023glyphcontrol, tuo2024anytext, liu2024glyph}. Subsequent efforts have expanded control over typography, stroke styles, multilingual rendering, and localized text erasure~\citep{zeng2024textctrl, liu2026scenevtg++, lu2026easytext}. Despite these advances, text synthesis in dynamic video remains largely unaddressed. While recent work such as SteerVTE~\citep{zeng2026steervte} begins to explore video text editing, the behavior of visual text in unified video generation remains under-characterized.

\subsection{Benchmarking Generative Video and Visual Text}
Current evaluation protocols remain divided between holistic video quality assessment and static image text benchmarks. In video generation, suites such as VBench~\citep{huang2024vbench} and EvalCrafter~\citep{liu2024evalcrafter} established standard metrics for visual realism, motion smoothness, and prompt adherence. Subsequent benchmarks expanded evaluation to intrinsic faithfulness, storytelling, and multi-shot generation~\citep{zheng2025vbench, wei2026msavbench, hua2026vabench, zeng2026beyond}. However, these benchmarks treat text as incidental scene content, failing to assess whether characters remain legible, stable, and editable over time. Conversely, dedicated visual text benchmarks remain predominantly image-based. Datasets such as AnyWord-3M~\citep{tuo2024anytext}, ScenePair~\citep{zeng2024textctrl}, and InnoText-Bench~\citep{liu2026innotext} evaluate sentence accuracy, edit distance, and typographic style similarity on paired static images. While effective for image-level evaluation, they cannot capture failure modes unique to dynamic video, such as temporal character decay, carrier detachment, and localized editing residue.

\section{Method}

\subsection{Data Design}
\label{sec:data-design}

\subsubsection{Data Construction and Task Routing}
\label{sec:data-construction}

VidScribe contains 803 human-verified samples derived from a task-agnostic metadata schema over 12 factor axes, organized into 3 core groups.
The \textit{Intrinsic Text Properties} group covers Text Amount (F1), Layout Hierarchy (F2), String Type (F3), Typographic Style (F4), Stroke Weight (F5), and Text Scale (F6).
The \textit{Physical Imaging Conditions} group covers Carrier Type (F7), Illumination (F8), and Text-Background Separability (F9).
The \textit{Temporal Behavior} group covers Motion Pattern (F10), Foreground Occlusion (F11), and Content Evolution (F12).
An automated feasibility filter eliminates physically contradictory combinations to ensure valid factor isolation before instantiation.
4 generation regimes share this unified factor space while exposing task-specific inputs.
Text-to-video~(T2V, \textbf{Write}) specifies all factors through text prompts alone.
Reference-to-video~(R2V, \textbf{Transfer}) provides intrinsic text appearance through an exemplar image together with scene and motion prompts.
Image-to-video~(I2V, \textbf{Sustain}) anchors initial text and scene geometry in a starting frame to test temporal persistence under dynamics.
Video-to-video~(V2V, \textbf{Edit}) supplies an existing video and requests localized text replacement, deletion, addition, translation, or restyling while preserving surrounding content.

\begin{figure*}[t]
    \centering
    \includegraphics[width=\textwidth]{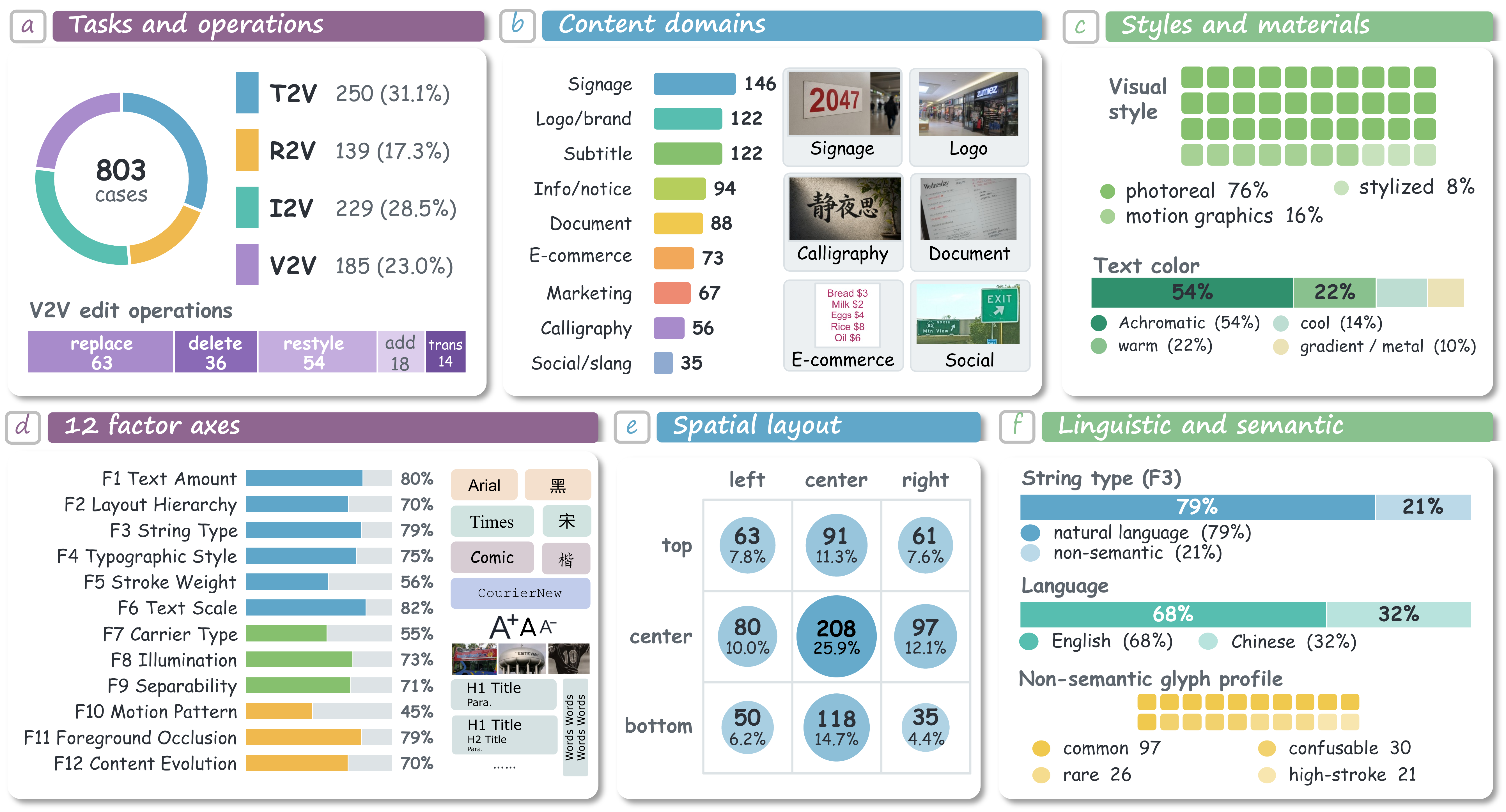}
    \caption{Data distribution of VidScribe across (a) generation regimes and V2V edit operations, (b) content domains, (c) visual styles and text materials, (d) the 12 factor axes, (e) spatial layout, and (f) linguistic and semantic properties.}
    \label{fig:data-distribution}
\end{figure*}

\subsubsection{Data Distribution and Taxonomic Analysis}
\label{sec:data-analysis}

VidScribe is designed to balance real-world scenario coverage, controlled generative stress, and intentional spatial-linguistic debiasing.

\textbf{Tasks and edit operations.}
As shown in Figure~\ref{fig:data-distribution}~(a), VidScribe distributes 803 samples across 4 generation regimes: T2V (250), R2V (139), I2V (229), and V2V (185). The V2V suite covers 5 distinct localized operations: replacement, deletion, restyling, addition, and translation.

\textbf{Content domains, styles, and materials.}
As illustrated in Figure~\ref{fig:data-distribution}~(b)--(c), VidScribe spans 9 real-world domains, balancing high-frequency text such as street signage, logos, and subtitles with demanding documents and complex calligraphy. The corpus covers photorealism, motion graphics, and stylized rendering. Text appearances span achromatic bases, warm tones, cool palettes, and specular metallic or gradient finishes.

\textbf{Factor coverage and spatial layout.}
To ensure controlled factor isolation, each of the 12 factor axes defines a default construction baseline, such as planar carriers or standard lighting.
As shown in Figure~\ref{fig:data-distribution}~(d), these baseline settings account for 45.0\% to 82.0\% across axes, ensuring sufficient standard generation contexts while systematically varying non-baseline attributes.
This design underpins our diagnostic factor analysis, which groups configurations into \textit{Regular} and \textit{Hard} regimes for damage ranking.
Furthermore, spatial coordinates follow a calibrated $3\times3$ grid in Figure~\ref{fig:data-distribution}~(e). While center placement accounts for 25.9\% of samples, the remaining 74.1\% occupy peripheral regions to prevent models from exploiting center-bias shortcuts.

\textbf{Linguistic and semantic properties.}
To decouple glyph rendering from language priors, Figure~\ref{fig:data-distribution}~(f) contrasts natural-language strings (79\%) with non-semantic sequences (21\%), such as randomized serial codes and license plates.
The benchmark maintains bilingual coverage, with 68\% English and 32\% Chinese samples.
The Chinese subset is stratified into common, visually confusable, rare, and high-stroke-density glyph profiles to benchmark fine-grained topological collapse.

\subsection{Evaluation Suite}
\label{sec:evaluation-suite}

\subsubsection{Evaluation Metrics}
\label{sec:metric-design}

VidScribe constructs a track-grounded spatio-temporal graph by associating region-restricted OCR detections~\citep{qwen37plus} across frames.
To prevent unresolvable evidence from inflating scores, strict measurability gating prunes frames where the text region leaves the field of view for a sustained interval, while algorithmic gates enforce block-level spelling vetoes, cascading score caps, and coverage thresholds (Appendix~\ref{sec:metric-computation}).
Under this grounding, we define 11 shared metrics and 2 task-specific probes across 5 groups.

The \textit{Text Fidelity} group measures lexical and structural correctness.
Content Accuracy (A1) compares consensus transcriptions against ground truth across visible frames, with a block-level veto on any spelling error.
Glyph Correctness (A2) inspects stroke topology across both target and generated background text, penalizing breaks, malformed loops, and structural anomalies.

The \textit{Temporal Stability} group evaluates persistence across frames.
Content Stability (B1) tracks character invariance across temporal windows, gated by content correctness.
Appearance Stability (B2) computes perceptual similarity across flow-aligned text patches between adjacent frames.

The \textit{Instruction Compliance} group audits prompt adherence on visual attributes.
Color Alignment (C1) evaluates visible stroke color against prompt specifications under scene lighting.
Stroke Weight Agreement (C2) scores normalized tier distance across ordered font weights (light, regular, bold).
Typographic Style Matching (C3) verifies font family correspondence across typeface categories.
Carrier Type Attachment (C4) checks blind categorization of the underlying carrier geometry.
Spatial Placement Accuracy (C5) checks target localization on a standardized $3 \times 3$ grid.
Layout Hierarchy Fidelity (C6) assesses multi-tier layout structure and relative line sizing.

The \textit{Non-Text Preservation} group evaluates background preservation. Background Consistency (D1) measures deep feature similarity across output frames for T2V, R2V, and I2V, and between aligned source and output frames for V2V, with text regions masked.

The \textit{Task-Specific Probes} group targets localized V2V editing failures.
Editing Residue (E1) detects obsolete source text persisting inside the edit region, while Source Text Preservation (E2) verifies that text outside the target region remains intact.

\subsubsection{Hierarchical Hybrid Evaluation Framework}
\label{sec:scoring-design}

\begin{figure*}[t]
    \centering
    \includegraphics[width=\textwidth]{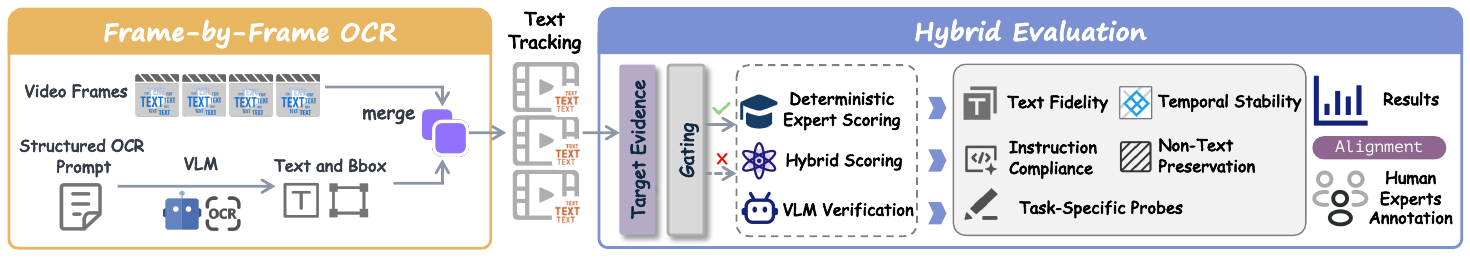}
    \caption{Overview of the VidScribe evaluation framework. Video frames are grounded through OCR and text tracking, then evaluated through 3 pathways: Deterministic Expert Scoring, Hybrid Scoring, and Rubric-Grounded VLM verification, with gating applied to enforce measurability.}
    \label{fig:framework}
\end{figure*}

Evaluating visual text in video requires heterogeneous evidence spanning stroke topology, physical motion, and layout semantics.
VidScribe therefore establishes a hierarchical hybrid evaluation framework that routes metrics through 3 specialized evidence pathways, as illustrated in Figure~\ref{fig:framework}.

\textbf{Path-I: Deterministic Expert Scoring.}
This pathway directly processes low-level physical, topological, and perceptual signals.
Glyph Correctness (A2) applies a structural inspector~\citep{textpecker} to identify stroke defects.
Appearance Stability (B2) and Background Consistency (D1) measure frame-to-frame fidelity via LPIPS~\citep{lpips} and CLIP features~\citep{clip}, while Stroke Weight Agreement (C2) and Typographic Style Matching (C3) use dedicated visual classifiers.

\textbf{Path-II: Hybrid Scoring with Expert Evidence and VLM Arbitration.}
This pathway resolves complex spatial attributes where raw geometric cues require semantic contextualization.
Specialized vision tools first extract objective geometric primitives, including surface normals, spatial bounding coordinates, and multi-line height ratios.
A vision-language arbiter~\citep{qwen37plus} then cross-references this structured evidence to verify higher-level physical adherence, including Carrier Type Attachment (C4), Spatial Placement Accuracy (C5), and Layout Hierarchy Fidelity (C6).

\textbf{Path-III: Rubric-Grounded VLM Verification.}
This pathway evaluates semantic fidelity, character strings, and discrete state transitions using prompt-specific rubrics with constrained categorical outputs.
It audits Content Accuracy (A1), Content Stability (B1), and Color Alignment (C1), and executes both V2V-specific probes, Editing Residue (E1) and Source Text Preservation (E2).

\subsection{Factor--Metric Mapping}
\label{sec:factor-metric-mapping}

VidScribe operates as a diagnostic instrument that links the 12 factor axes to generative performance.
To establish an interpretable diagnostic correspondence, we define a design map that pairs each factor axis with its primary evaluation metric across 2 assignment tiers (Appendix~\ref{app:axis-metric-correspondence}).

\textbf{Definitional assignments.}
For axes whose generation target is explicitly evaluated, the assigned metric directly restates the axis itself:
Layout Hierarchy (F2) maps to Layout Hierarchy Fidelity (C6), Typographic Style (F4) maps to Typographic Style Matching (C3), Stroke Weight (F5) maps to Stroke Weight Agreement (C2), and Carrier Type (F7) maps to Carrier Type Attachment (C4).

\textbf{Mechanism assignments.}
For axes that introduce physical, structural, or temporal perturbations, the assigned metric is determined by the primary degradation mechanism where failures first emerge:
Text Amount (F1) maps to Content Accuracy (A1), as heavier character loads increase lexical omissions;
String Type (F3) and Foreground Occlusion (F11) map to Glyph Correctness (A2), as non-semantic sequences and physical occlusions directly stress visual stroke topology by stripping language priors or breaking contours;
Content Evolution (F12) maps to Content Stability (B1), directly challenging sequence persistence;
Illumination (F8) and Text-Background Separability (F9) map to Color Alignment (C1), where adverse lighting, reflections, and low contrast induce apparent color bleeding.
Text Scale (F6) and Motion Pattern (F10) map to Spatial Placement Accuracy (C5), as tiny fonts and carrier motion tighten and disrupt spatial bounding tolerances.

\section{Experiments}
\label{sec:experiments}

We benchmark 7 commercial video generators on VidScribe: MiniMax H3~\citep{minimax2026h3}, Seedance 2.5~\citep{seedance202625}, Wan 3.0~\citep{wan2026wan27}, Gemini Omni 1.1 Flash~\citep{geminiomni}, HappyHorse~\citep{happyhorse2026}, PixVerse V6~\citep{PixVerse}, and Kling3.0-Omni~\citep{kuaishou2024kling}. We also evaluate 4 specialized open-source models: HunyuanVideo-1.5~\citep{wu2025hunyuanvideo} and LTX-2.5~\citep{hacohen2026ltx} for T2V, and JoyAI-Video-Edit~\citep{xiao2026joyai} and Kiwi-Edit~\citep{lin2026kiwi} for V2V, with results in Appendix~\ref{app:task-wise-model-performance}. Unless stated otherwise, all metric scores are reported as percentages (\%), where higher is better except for Editing Residue (E1).

\subsection{Main Results}
\label{sec:experiments_main}

\begin{table}[h]
\centering
\footnotesize
\setlength{\tabcolsep}{2.0pt}
\caption{Results of 7 commercial unified video generators on VidScribe. ``Overall'' denotes the group-balanced mean over the 11 shared metrics.}
\label{tab:closemain}
\begin{tabular}{l c cc cc cccccc c}
\toprule
& & \multicolumn{2}{c}{\shortstack{\textit{Text}\\\textit{Fidelity}}}
  & \multicolumn{2}{c}{\shortstack{\textit{Temporal}\\\textit{Stability}}}
  & \multicolumn{6}{c}{\shortstack{\textit{Instruction}\\\textit{Compliance}}}
  & \multicolumn{1}{c}{\shortstack{\textit{Non-Text}\\\textit{Preservation}}} \\
\cmidrule(lr){3-4} \cmidrule(lr){5-6} \cmidrule(lr){7-12} \cmidrule(lr){13-13}
Video generators & Overall $\uparrow$ & A1 $\uparrow$ & A2 $\uparrow$ & B1 $\uparrow$ & B2 $\uparrow$
& C1 $\uparrow$ & C2 $\uparrow$ & C3 $\uparrow$ & C4 $\uparrow$ & C5 $\uparrow$ & C6 $\uparrow$
& D1 $\uparrow$ \\
\midrule
MiniMax H3 & \best{79.2} & \second{82.1} & \second{57.0} & \second{74.8} & 83.5
& 83.5 & \second{74.3} & 75.3 & \second{75.2} & 74.1 & 83.4 & \second{90.6} \\
Wan 3.0 & \second{78.5} & \best{82.7} & 45.4 & \best{76.7} & \best{86.0}
& \second{84.4} & \best{75.6} & \best{79.7} & \best{77.7} & \second{76.8} & \best{83.9} & 88.8 \\
Seedance 2.5 & 78.1 & 77.6 & \best{58.2} & 72.5 & 83.4
& \best{84.5} & 72.9 & 70.9 & 75.0 & 71.1 & 80.9 & \best{90.7} \\
Gemini Omni 1.1 Flash & 77.8 & 79.0 & 49.9 & 73.8 & \second{84.4}
& 82.6 & 72.9 & \second{76.6} & 73.1 & \best{77.1} & \second{83.7} & 90.1 \\
HappyHorse & 75.0 & 72.5 & 48.7 & 66.7 & 82.5
& 84.1 & 72.7 & 70.3 & 73.5 & 71.0 & 82.2 & 89.1 \\
PixVerse V6 & 73.2 & 71.6 & 48.1 & 62.7 & 78.4
& 83.2 & 73.9 & 75.8 & 68.9 & 65.8 & 82.9 & 87.3 \\
Kling3.0-Omni & 70.5 & 57.3 & 43.8 & 54.1 & 80.7
& 81.4 & 73.7 & 67.5 & 72.5 & 69.8 & 82.9 & 89.6 \\
\bottomrule
\end{tabular}
\end{table}

\begin{table*}[h]
\centering

\begin{minipage}[t]{0.54\textwidth}
\vspace{0pt}
\centering

\caption{Task-wise performance and V2V-specific editing diagnostics across unified video generators.}
\label{tab:task_and_v2v}
\begingroup
\scriptsize
\setlength{\tabcolsep}{2.8pt}
\renewcommand{\arraystretch}{1.08}

\begin{adjustbox}{max width=\linewidth}
\begin{tabular}{@{}l cccc cc@{}}
\toprule
& \multicolumn{4}{c}{\textbf{Task-wise overall} $\uparrow$}
& \multicolumn{2}{c}{\textbf{V2V probes}} \\
\cmidrule(lr){2-5}
\cmidrule(l){6-7}
Model
& T2V & R2V & I2V & V2V
& \shortstack{Residue\\E1 $\downarrow$}
& \shortstack{Preserve\\E2 $\uparrow$} \\
\midrule
MiniMax H3
& 80.2 & \textbf{81.9} & \textbf{84.2} & 68.7
& 29.2 & 49.6 \\
Wan 3.0
& 79.6 & 80.4 & 81.8 & \textbf{70.3}
& 19.5 & \textbf{69.5} \\
Seedance 2.5
& 80.2 & 80.9 & 83.5 & 64.9
& 34.5 & 62.7 \\
Gemini Omni 1.1 Flash
& \textbf{80.9} & 76.3 & 81.6 & 69.3
& 28.3 & 68.3 \\
HappyHorse
& 76.6 & 78.3 & 80.4 & 62.3
& 29.2 & 38.5 \\
PixVerse V6
& 71.0 & 75.4 & 80.5 & 64.6
& \textbf{18.6} & 33.1 \\
Kling3.0-Omni
& 65.9 & 71.1 & 83.5 & 58.9
& 39.8 & 55.0 \\
\midrule
\textit{Task mean}
& 76.3 & 77.8 & \textbf{82.2} & 65.6
& 28.4 & 53.8 \\
\bottomrule
\end{tabular}
\end{adjustbox}
 
\endgroup
\end{minipage}%
\hfill
\begin{minipage}[t]{0.40\textwidth}
\vspace{0pt}
\centering

\includegraphics[
  width=\linewidth
]{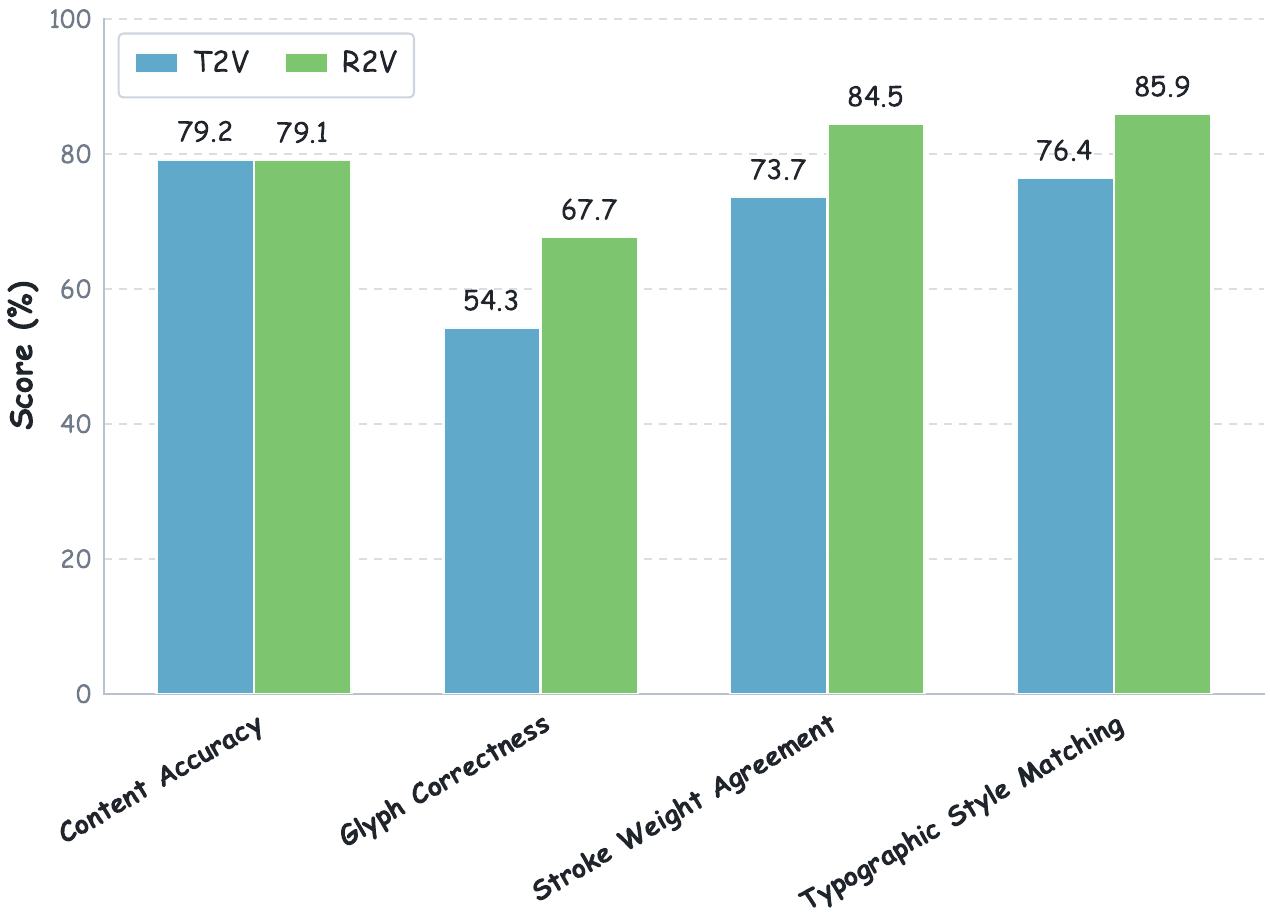}

\captionof{figure}{
Controlled T2V--R2V comparison under matched contexts.
}
\label{fig:t2v_vs_r2v}

\end{minipage}

\end{table*}

\textbf{Finding 1: Video text capability is non-monolithic, with a clear split between lexical correctness and glyph fidelity.}
Table~\ref{tab:closemain} shows that no model dominates across all dimensions. MiniMax H3 achieves the highest Overall score (79.2\%), yet does not lead any individual metric. Wan 3.0 attains the highest Content Accuracy (A1: 82.7\%) and the strongest Temporal Stability (B1: 76.7\%, B2: 86.0\%), but drops sharply on Glyph Correctness (A2: 45.4\%), trailing Seedance 2.5 (58.2\%) by 12.8\%. Across all systems, A1 reaches 82.7\%, while A2 remains below 60.0\%. This gap shows that correct character identity often coexists with broken stroke topology. The same pattern appears in instruction compliance: color and layout hierarchy remain relatively robust, while stroke weight and font style are consistently weaker. Visual text in video is therefore not a single capability, but a combination of lexical, structural, and typographic skills that degrade differently.

\textbf{Finding 2: Performance is strongly task-asymmetric, with I2V as the most reliable regime and V2V as the primary bottleneck.}
Table~\ref{tab:task_and_v2v} reveals a consistent ordering across regimes. I2V yields the highest task mean (82.2\%), showing that current unified video generators are comparatively strong at sustaining already instantiated text under different motion patterns. V2V is substantially harder, with the task mean dropping to 65.6\%, a 16.6\% gap relative to I2V. Even the strongest V2V models, Wan 3.0 and MiniMax H3, reach only 70.3\% and 68.7\%, while Kling3.0-Omni falls to 58.9\%. Between open-ended generation settings, R2V outperforms T2V for 6 of 7 systems, raising the task mean from 76.3\% to 77.8\%. This gain is not universal, as Gemini Omni 1.1 Flash performs better on T2V than R2V. Overall, the benchmark reveals a stable capability ordering: preserving text is easier than writing it, and localized rewriting is hardest.

\begin{table*}[t]
\centering
\footnotesize
\setlength{\tabcolsep}{4.5pt}
\renewcommand{\arraystretch}{0.97}
\caption{Factor sensitivity under \textit{Regular} and \textit{Hard} settings. Factors are ordered by task-balanced Overall Cliff's $\delta$ (\textit{Hard} vs. \textit{Regular}), where more negative values indicate larger degradation under \textit{Hard}. \textit{Regular/Hard} definitions are given in Appendix~\ref{app:metadata-schema}.}
\label{tab:factor_sensitivity}
\begin{tabular}{l lccc ccc}
\toprule
 & \multicolumn{4}{c}{\textbf{Assigned Metric}} & \multicolumn{3}{c}{\textbf{Overall Score}} \\
\cmidrule(lr){2-5} \cmidrule(lr){6-8}
\textbf{Factor Axis} & Metric & \textit{Regular} & \textit{Hard} & $\Delta$ & \textit{Regular} & \textit{Hard} & $\Delta$ \\
\midrule
F12: Content Evolution & B1: Content Stability & 70.8 & 49.7 & $-21.1$ & 77.3 & 69.4 & $-7.9$ \\
F11: Foreground Occlusion & A2: Glyph Correctness & 52.5 & 37.4 & $-15.1$ & 76.8 & 71.8 & $-5.0$ \\
F2: Layout Hierarchy & C6: Hierarchy Fidelity & 87.7 & 50.9 & $-36.8$ & 76.7 & 72.6 & $-4.1$ \\
F10: Motion Pattern & C5: Placement Accuracy & 75.0 & 67.9 & $-7.1$ & 77.0 & 73.9 & $-3.1$ \\
F1: Text Amount & A1: Content Accuracy & 74.7 & 66.0 & $-8.7$ & 76.7 & 73.6 & $-3.1$ \\
F6: Text Scale & C5: Placement Accuracy & 73.0 & 66.3 & $-6.7$ & 76.7 & 73.1 & $-3.6$ \\
F5: Stroke Weight & C2: Weight Agreement & 81.2 & 65.6 & $-15.6$ & 77.3 & 74.3 & $-3.0$ \\
F4: Typographic Style & C3: Style Matching & 80.5 & 56.6 & $-23.9$ & 76.8 & 74.5 & $-2.3$ \\
F9: Separability & C1: Color Alignment & 82.9 & 79.7 & $-3.2$ & 76.4 & 74.8 & $-1.6$ \\
F8: Illumination & C1: Color Alignment & 82.6 & 75.8 & $-6.8$ & 76.1 & 75.3 & $-0.8$ \\
F3: String Type & A2: Glyph Correctness & 48.6 & 47.9 & $-0.7$ & 76.0 & 75.4 & $-0.6$ \\
F7: Carrier Type & C4: Carrier Attachment & 80.3 & 61.4 & $-18.9$ & 76.3 & 75.6 & $-0.7$ \\
\bottomrule
\end{tabular}
\end{table*}

\textbf{Finding 3: Performance degradation is more concentrated on a small subset of text-centric structural and temporal factors than on adverse imaging conditions.}
Table~\ref{tab:factor_sensitivity} shows that the largest Overall drops come from Content Evolution, Foreground Occlusion, and Layout Hierarchy, reducing Overall by 7.9\%, 5.0\%, and 4.1\%, respectively. Motion Pattern, Text Scale, and Text Amount form the next tier. Imaging factors are much weaker: Separability and Illumination reduce Overall by only 1.6\% and 0.8\%, while Carrier Type and String Type reduce it by 0.7\% and 0.6\%. Assigned-metric drops isolate a factor's most direct failure mode, while the Overall ranking reflects whether that failure propagates across dimensions. Typographic Style and Carrier Type lose 23.9\% and 18.9\% on their assigned metrics, but only 2.3\% and 0.7\% on Overall, indicating localized degradation. In contrast, the highest-ranked structural and temporal factors degrade multiple metric groups together. Full factor--metric analysis is reported in Appendix~\ref{app:axis-metric-correspondence}.

\subsection{Further Diagnostic Analysis}
\label{sec:further-analysis}

\begin{figure}[t]
    \centering
    \includegraphics[width=0.9\linewidth]{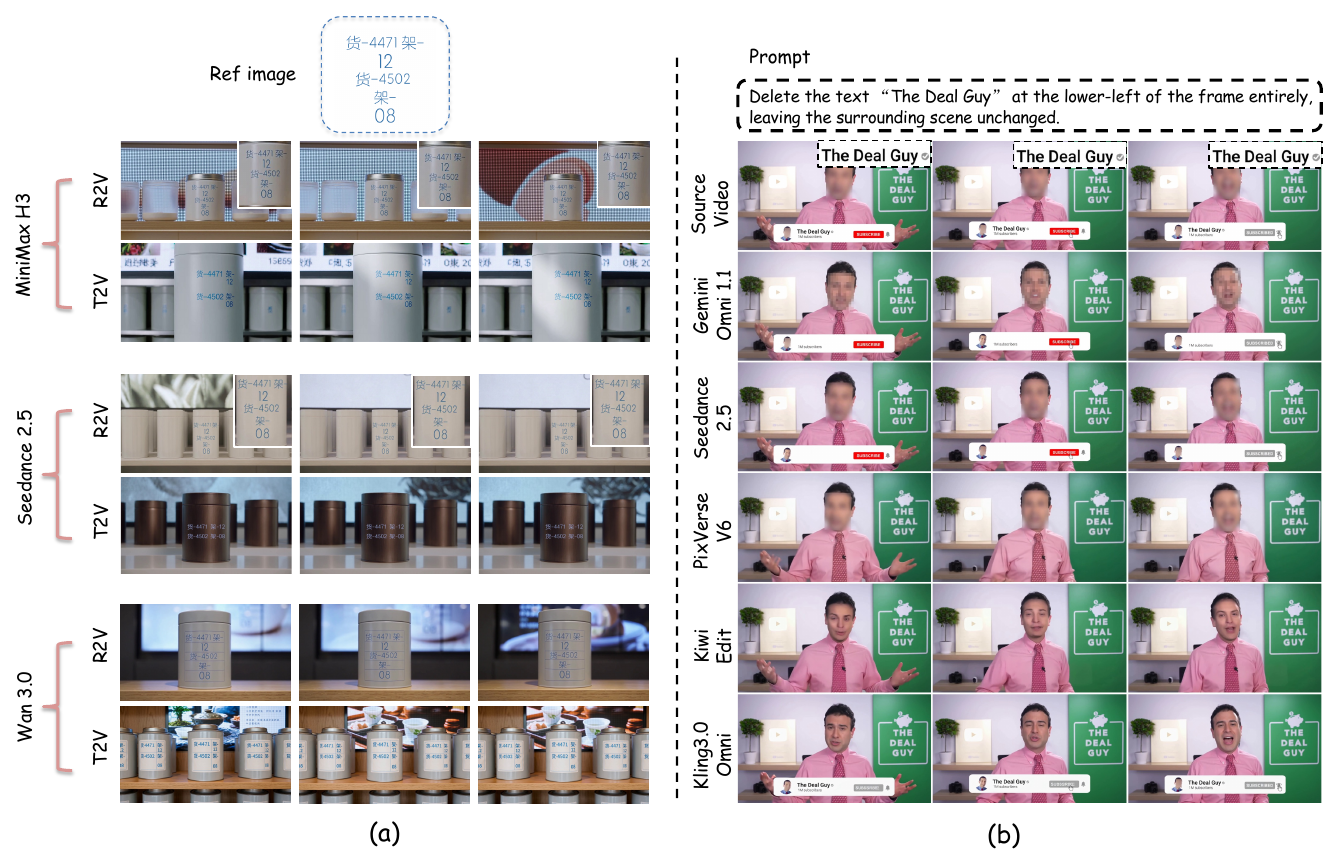}
    \caption{Further diagnostic examples. (a) Matched R2V--T2V comparisons. (b) A representative V2V failure case with over-erasure or residual source text.}
    \label{fig:r2v_v2v_case}
\end{figure}

\textbf{Visual reference enhances glyph and typographic fidelity rather than content accuracy.}
Under matched contexts in Figure~\ref{fig:t2v_vs_r2v}, supplying an exemplar image (R2V) yields no improvement in spelling accuracy alone.
By contrast, visual reference drives substantial gains in structural realization: Glyph Correctness on non-semantic strings (A2) surges by $\mathbf{+13.4\%}$ (54.3\% to 67.7\%), Stroke Weight Agreement (C2) gains $\mathbf{+10.8\%}$, and Typographic Style Matching (C3) increases by $\mathbf{+9.5\%}$ (Figure~\ref{fig:r2v_v2v_case}~(a)).
Visual exemplars effectively anchor glyph topology, but do not alleviate spelling errors.

\textbf{Localized V2V editing exposes limitations in fine-grained text perception and target isolation.}
In Table~\ref{tab:task_and_v2v}, the \textit{Task-Specific Probes} evaluate obsolete Editing Residue (E1) and Source Text Preservation (E2). Low residue does not imply strong preservation. PixVerse V6 and Wan 3.0 achieve similarly low residue (18.6\% vs. 19.5\%), yet Wan 3.0 preserves more than twice as much surrounding source text (69.5\% vs. 33.1\%). As shown in Figure~\ref{fig:r2v_v2v_case}~(b), even top systems either leave obsolete character fragments inside the edit region or alter valid adjacent text outside it. The main difficulty of V2V is therefore isolating the target region without leaking edits into nearby content.

\textbf{VidScribe as an actionable training signal.}
Beyond diagnostic evaluation, VidScribe provides verifiable signals for model alignment.
To prevent evaluation contamination, we construct an additional independent set of 1{,}000 prompts derived from our schema to build preference pairs and fine-tune Wan2.2-TI2V-5B~\citep{wan2025wan} via Direct Preference Optimization (DPO)~\citep{wallace2024diffusion}.
Evaluated on the held-out VidScribe T2V benchmark, aligned DPO improves the overall mean from 45.9\% to 48.4\% ($+2.5\%$), with gains concentrated in key structural bottlenecks: Layout Hierarchy (C6: $+9.3\%$), Glyph Correctness (A2: $+5.0\%$), Spatial Placement Accuracy (C5: $+4.6\%$), and Content Accuracy (A1: $+4.2\%$). Training details are shown in Appendix~\ref{app:dpo}.

\subsection{Human Preference Alignment and Evaluation Robustness}
\label{sec:human-robustness}

\begin{table}[t]
\centering
\footnotesize
\setlength{\tabcolsep}{2.8pt}
\caption{Correlation with blinded human expert consensus on 950 sampled videos ($\rho$ and $r \in [-1, 1]$, all $p < 0.001$). Qwen3.7-Plus serves as our default evaluator for reporting results.}
\label{tab:human_align}
\begin{tabular}{ll ccccccccccc}
\toprule
Evaluator Backbone & Correlation & A1 & A2 & B1 & B2 & C1 & C2 & C3 & C4 & C5 & C6 & D1 \\
\midrule
\multirow{2}{*}{\textbf{Qwen3.7-Plus}} 
& Spearman ($\rho$) $\uparrow$ & 0.87 & 0.84 & 0.82 & 0.81 & 0.78 & 0.84 & 0.91 & 0.72 & 0.88 & 0.79 & 0.96 \\
& Pearson ($r$) $\uparrow$    & 0.90 & 0.90 & 0.85 & 0.83 & 0.77 & 0.86 & 0.86 & 0.72 & 0.89 & 0.76 & 0.94 \\
\midrule
\multirow{2}{*}{\textbf{Gemini-3.8-Flash}} 
& Spearman ($\rho$) $\uparrow$ & 0.82 & 0.80 & 0.83 & 0.79 & 0.76 & 0.84 & 0.87 & 0.70 & 0.82 & 0.79 & 0.95 \\
& Pearson ($r$) $\uparrow$    & 0.85 & 0.84 & 0.85 & 0.80 & 0.76 & 0.85 & 0.86 & 0.71 & 0.85 & 0.77 & 0.94 \\
\bottomrule
\end{tabular}
\end{table}

To validate evaluation reliability, 8 domain experts in typography and video production rated 950 videos sampled uniformly from all 11 systems, with each video evaluated by at least 5 experts. Table~\ref{tab:human_align} shows strong alignment between VidScribe and expert judgment across all 11 shared metrics. Under the default Qwen3.7-Plus evaluator, Glyph Correctness (A2) and Typographic Style Matching (C3) reach Spearman correlations of 0.84 and 0.91, while both \textit{Temporal Stability} metrics exceed 0.80. Background Consistency (D1) reaches 0.96, confirming that the benchmark captures both localized text failures and collateral scene corruption. Replacing Qwen3.7-Plus with Gemini-3.8-Flash yields closely matched correlations, showing that VidScribe is robust to the choice of evaluator backbone.

\section{Conclusion}

We present \textbf{VidScribe}, a unified benchmark for visual text in video across T2V, R2V, I2V, and V2V. By combining a 12-axis factor space with a track-grounded evaluation suite, VidScribe enables fine-grained evaluation of writing, transfer, sustaining, and localized editing under diverse physical-imaging and temporal conditions. Experiments show that visual text capability is non-monolithic and strongly task-asymmetric. Current video generation systems often maintain recognizable content while failing on glyph correctness, perform far more reliably in I2V than in V2V editing, and degrade more under text-centric structural and temporal dynamics than under adverse imaging conditions. We further show that visual reference mainly improves glyph and typographic fidelity rather than content accuracy, while localized editing remains limited by incomplete source-text removal and weak preservation of undeclared text. We hope VidScribe serves as a reliable diagnostic benchmark and supports future progress in video text generation and editing.

\section*{AI Use Statement}

In this work, we used generative AI tools for generating prompts and cleaning the prompt dataset, which have been claimed in the main text, and also assisting in the writing of proofs and assisting with translation.
We have not used generative AI tools for helping develop theoretical models or conceptual frameworks, designing or providing feedback on research methodology or experiments, implementing methods, supporting qualitative and thematic data analysis or interpreting results.
Formulating mathematical claims, providing critical ingredients for proving mathematical claims, proposing or refining hypotheses are not applicable to this work.
We have reviewed all AI-assisted work. We take responsibility for the final content of this work, including text, claims or artifacts produced with the aid of generative AI.



\section*{Ethics Statement}

The text prompts used in VidScribe are generated by LLMs, subject to constraints, and have undergone human review. All collected or generated images and videos have been screened and manually vetted to exclude real human faces. These prompts, images, and videos contain no sensitive information, political content, violence, pornography, or other prohibited material, nor do they involve real names or other personally identifiable information. Furthermore, the images and videos generated by VidScribe are intended solely for benchmarking purposes and are not used for any other applications.

\bibliography{iclr2027_conference}
\bibliographystyle{iclr2027_conference}

\appendix
\clearpage

\section{Dataset Details}
\label{app:dataset-details}

\subsection{Shared Metadata Schema and Sample Records}
\label{app:metadata-schema}

VidScribe comprises 803 human-verified samples, each instantiated as a nominal 5-second single-shot clip. Most samples use a landscape format at $1280\times720$, while the V2V split additionally includes a portrait subset at $720\times1280$ to cover mobile-style editing scenarios. Each sample is stored as a single JSON object, which we refer to as a \emph{sample metadata record}. The shared \emph{metadata schema} specifies the fields, value vocabularies, representations, and conditional applicability rules for these records. Each record contains the task label, the 12 factor axes, coverage attributes, conditional controls, task-specific ground-truth strings, and the prompt with optional visual input. Non-applicable fields are stored as \texttt{null}. For V2V, \texttt{gt\_text} denotes the source text and \texttt{gt\_text\_after} denotes the edited target. For temporal content switching, \texttt{gt\_text\_switched} stores the post-switch target. For off-screen cropping, \texttt{gt\_text\_visible} stores the visible portion used for evaluation.

To preserve compatibility with the released annotations and evaluation code, JSON keys retain stable implementation-facing names, even when they differ from the display names used in the paper. For the 12 factor axes, the factor indices \textbf{F1--F12} serve as the canonical identifiers linking each JSON field to its corresponding paper definition.

The schema supports both scalar and level-aware representations. In multi-level records, each level-sensitive field is stored as an ordered list of \texttt{\{"level": $i$, "value": $v_i$\}} entries, even when the same value is shared across levels; sample-level fields remain scalar. Multi-instance targets are stored as arrays. Aggregate statistics use the first target or the level-1 value as the sample-level representative, while the original representation is preserved for level-aware metrics and multi-instance analysis.

We next show one illustrative sample metadata record and two partial record fragments. The first shows how a T2V record is organized; the fragments separately illustrate level-aware attributes and multi-instance targets.

\begin{tcolorbox}[
  enhanced,breakable,
  colback=gray!5,colframe=gray!55!black,
  arc=2pt,boxrule=0.45pt,
  left=5pt,right=5pt,top=4pt,bottom=4pt,
  fonttitle=\bfseries\small,
  title={Illustrative sample metadata record (T2V; prompt abridged)}
]
\scriptsize
\begin{verbatim}
{
  "sample_id": "TRE-T2V-RL3-243",
  "task": "T2V",
  "F1_text_amount": "moderate",
  "F2_hierarchy": "single_level",
  "F3_string_type": "natural_language",
  "F4_typography_style": "serif",
  "F5_typography_weight": "regular",
  "F6_text_scale": "medium",
  "F7_carrier": "flat_2d",
  "F8_lighting": "normal",
  "F9_text_bg_separability": "high",
  "F10_spatiotemporal_behavior": "camera_motion_world_locked",
  "F11_occlusion": "none",
  "F12_content_dynamics": "stable",
  "language": "en",
  "content_domain": "ecommerce",
  "visual_style": "photoreal",
  "text_color": "gradient_or_metallic",
  "v2v_edit_op": null,
  "font_name": "TimesNewRoman",
  "text_color_desc": "metallic_silver",
  "motion_pattern": "camera_dolly_right",
  "occluder_type": null,
  "deform_source": null,
  "text_position": "center_right",
  "gt_text": "Same Price",
  "gt_text_after": null,
  "gt_text_switched": null,
  "gt_text_visible": null,
  "glyph_profile": null,
  "prompt": "A photorealistic gift shop counter ...",
  "reference_image": null,
  "source_video": null,
  "temporal_evidence": ["F10_camera_motion_world_locked"]
}
\end{verbatim}
\end{tcolorbox}

\begin{tcolorbox}[
  enhanced,breakable,
  colback=gray!5,colframe=gray!55!black,
  arc=2pt,boxrule=0.45pt,
  left=5pt,right=5pt,top=4pt,bottom=4pt,
  fonttitle=\bfseries\small,
  title={Level-aware and multi-instance record fragments}
]
\scriptsize
\begin{verbatim}
[
  {
    "sample_id": "TRE-T2V-0005",
    "F2_hierarchy": "multi_level_plus",
    "F6_text_scale": [
      {"level": 1, "value": "large"},
      {"level": 2, "value": "medium"},
      {"level": 3, "value": "tiny"}
    ],
    "gt_text": [
      {"level": 1, "value": "New Sale"},
      {"level": 2, "value": "Deal Now"},
      {"level": 3, "value": "Ends Soon"}
    ]
  },
  {
    "sample_id": "TRE-T2V-r2-028",
    "F1_text_amount": "multi_instance",
    "F2_hierarchy": "single_level",
    "gt_text": [
      "SALE", "NEW ARRIVAL", "LIMITED",
      "SHOP NOW", "BUY MORE SAVE MORE"
    ]
  }
]
\end{verbatim}
\end{tcolorbox}

Table~\ref{tab:app-twelve-axes} summarizes the 12-axis factor space, including the baseline value for each axis, its non-baseline values, and the corresponding operational distinction. For factor sensitivity analysis, underlined values are treated as \textit{Hard}, and all remaining values as \textit{Regular}. Multi-level samples are labeled \textit{Hard} if any level is \textit{Hard}. Before grouping F8, legacy \texttt{low\_key} and \texttt{backlit} labels are normalized to \texttt{extreme}.

Beyond the 12 factor axes, the schema also includes auxiliary fields for coverage and diagnosis. Language, content domain, visual style, and text color are coverage attributes rather than factor axes. For Chinese scoring text, \texttt{glyph\_profile} stores one mutually exclusive difficulty category; it is not used for V2V. \texttt{font\_name} and \texttt{text\_color\_desc} record the concrete typeface and color realization. F10 specifies the coarse motion-and-anchoring class, whereas \texttt{motion\_pattern} specifies the concrete camera or text motion when applicable. The remaining control fields are \texttt{occluder\_type}, \texttt{deform\_source}, and \texttt{text\_position}. \texttt{temporal\_evidence} records observable dynamic cues, while \texttt{reference\_image} and \texttt{source\_video} route task-specific visual inputs; non-applicable fields are \texttt{null}. V2V additionally uses \texttt{v2v\_edit\_op} to mark the localized edit type.

\begin{table*}[t]
\centering
\caption{VidScribe's 12 factor axes and value definitions. Bold marks baseline values; underlining marks \textit{Hard} values, with all remaining values treated as \textit{Regular}.}
\label{tab:app-twelve-axes}

\scriptsize\rmfamily
\setlength{\tabcolsep}{3.8pt}
\renewcommand{\arraystretch}{1.12}

\newlength{\vsnbcol}%
\settowidth{\vsnbcol}{\scriptsize\rmfamily camera\_motion\_screen\_locked}%
\addtolength{\vsnbcol}{1.5pt}%
\newlength{\vsdefcol}%
\setlength{\vsdefcol}{\dimexpr0.580\textwidth-\vsnbcol\relax}%

\begin{tabular}{@{}
p{0.040\textwidth}
p{0.120\textwidth}
p{0.135\textwidth}
p{\vsnbcol}
p{\vsdefcol}
@{}}
\toprule
\textbf{Axis} &
\textbf{Factor} &
\textbf{Baseline} &
\textbf{Non-baseline values} &
\textbf{Definition} \\
\midrule

\multicolumn{5}{@{}l}{\textit{Intrinsic text properties} (\textbf{F1--F6})} \\
\addlinespace[2pt]

F1 &
Text Amount &
\textbf{minimal} &
\raggedright moderate, \underline{heavy}, \underline{multi\_instance} &
Character load and instance count in one frame: 1--8, 9--30, or 31+
characters, or 5+ separate blocks. \\

F2 &
Layout Hierarchy &
\textbf{single\_level} &
\raggedright two\_level, \underline{multi\_level\_plus}, \underline{table\_like} &
Number of typographic levels, from single level to title--body,
multi-level, or tabular. \\

F3 &
String Type &
\textbf{natural\_language} &
\raggedright \underline{non\_semantic} &
Whether the string follows natural language or is a non-semantic code
such as a serial number or plate. \\

F4 &
Typographic Style &
\textbf{sans} &
\raggedright serif, \underline{script}, \underline{monospace} &
Coarse typeface category. \\

F5 &
Stroke Weight &
\textbf{regular} &
\raggedright \underline{light}, \underline{bold} &
Ordered stroke-weight class. \\

F6 &
Text Scale &
\textbf{medium} &
\raggedright large, \underline{tiny} &
Single-line height at 720p: 22--108\,px, above 108\,px, or at most
21\,px. \\

\midrule
\multicolumn{5}{@{}l}{\textit{Physical imaging conditions} (\textbf{F7--F9})} \\
\addlinespace[2pt]

F7 &
Carrier Type &
\textbf{flat\_2d} &
\raggedright \underline{3d\_surface}, freestanding, \underline{handwritten} &
How the text is supported, overlaid, or inscribed in the scene. \\

F8 &
Illumination &
\textbf{normal} &
\raggedright \underline{challenging}, \underline{extreme} &
Lighting difficulty, from normal to extreme. \\

F9 &
Text-Background Separability &
\textbf{high} &
\raggedright \underline{low}, \underline{busy\_texture} &
How reliably text strokes separate from the local background in
luminance, hue, or texture. \\

\midrule
\multicolumn{5}{@{}l}{\textit{Temporal behavior} (\textbf{F10--F12})} \\
\addlinespace[2pt]

F10 &
Motion Pattern &
\textbf{static} &
\raggedright\hyphenpenalty=10000\exhyphenpenalty=10000
\underline{text\_motion\_screen}, \newline
camera\_motion\_screen\_locked, \newline
\underline{camera\_motion\_world\_locked}, \newline
\underline{camera\_motion\_deformable} &
Text or camera motion relative to screen, world, or deformable-scene
coordinates. \\

F11 &
Foreground Occlusion &
\textbf{none} &
\raggedright partial, \underline{persistent}, \underline{intermittent},
\underline{off\_screen\_crop} &
Target-text visibility pattern under occlusion. \\

F12 &
Content Evolution &
\textbf{stable} &
\raggedright \underline{content\_switch}, \underline{scroll}, \underline{typing},
\underline{handwriting\_stroke} &
Whether the text content itself changes or is progressively revealed
over time, as opposed to positional motion (F10). \\

\bottomrule
\end{tabular}
\end{table*}

\subsection{Task-Specific Construction and Prompts}
\label{app:data-construction}

Building on the task routing in Section~\ref{sec:data-construction}, this section details how each task is instantiated and pairs it with a representative prompt. All prompts are written in English, while quoted target strings remain in their original language. Prompts expose only model-facing conditions and omit evaluator-side metadata. Temporal instructions specify the moving entity, direction, magnitude, anchoring rule, and time span. The initial text state must be visible from the first frame, so typing and handwriting samples begin from the first visible character or stroke. When F10, F11, and F12 are all at baseline, the prompt instead includes visible scene motion with a static camera.

\paragraph{T2V.}
Each T2V sample begins with a content domain, target string, factor configuration, and the corresponding coverage attributes. Deterministic checks remove unsupported tokens, missing fields, inconsistent ground truth, and infeasible combinations. The accepted sample specifications are then converted into English prompts with a slot-based generator, followed by human review for exact target strings, field--prompt consistency, and instruction validity.

\begin{tcolorbox}[
  enhanced,breakable,
  colback=blue!3,colframe=blue!45!black,
  arc=2pt,boxrule=0.45pt,
  left=5pt,right=5pt,top=5pt,bottom=5pt,
  fonttitle=\bfseries\small,
  title={T2V Prompt Exemplar (Writing from language)}
]
\footnotesize
A photorealistic city street shot at dusk. From the first frame, English sign text appears on a flat wall-mounted signboard, positioned on the right side of the frame at mid-height. The text is set in yellow TeX Gyre Bonum serif letters with regular weight and medium size. It is a single-level text element and reads ``Harbor Road.'' The lighting is uneven, with reflections and shadows across the text. The text has high contrast against the background. The camera dollies steadily to the right at a constant speed, sweeping about one third of the frame width over the clip. The text remains rigidly attached to its real-world surface, undergoing the correct perspective and scale change as the viewpoint shifts, with no drift relative to the surface. There is no occlusion. The text stays fully visible for the entire clip. The text content remains unchanged.
\end{tcolorbox}

\paragraph{R2V.}
For R2V, we render a clean $1024\times1024$ text reference using one of 11 supported real fonts. The reference provides text identity and intrinsic appearance, including typographic style, stroke weight, and color, but not scene layout or temporal behavior. Long strings are packed into centered rows, and reference renders below the minimum line-height threshold are regenerated. For Chinese samples, we verify font support for every scored glyph.
In the 803-sample benchmark, the R2V split contains 139 samples. In this split, the prompt explicitly includes the target text so that text identity is specified in language, while the reference image supplies the corresponding visual form, including glyph shape, typographic style, stroke weight, and color. This setting defines the main R2V benchmark condition used throughout the dataset.

Figure~\ref{fig:t2v_vs_r2v} and figure~\ref{fig:r2v_v2v_case} are based on a separate controlled comparison derived from these 139 R2V samples. For each sample, we keep the scene specification, motion pattern, layout, target text instance, and annotated factor settings unchanged, and construct an auxiliary R2V prompt in which the \texttt{gt\_text} mention is removed, while the paired reference image remains the same. We then build a matched T2V counterpart under the same scenario configuration. Unlike the auxiliary R2V setting, this T2V prompt contains the complete text and scene specification required for text-only generation and does not use any reference image. The comparison in Figure~\ref{fig:t2v_vs_r2v} therefore aligns T2V and R2V at the scenario level, so the results reported in the main text are grounded in matched generation conditions rather than scene mismatch.

\begin{tcolorbox}[
  enhanced,breakable,
  colback=blue!3,colframe=blue!45!black,
  arc=2pt,boxrule=0.45pt,
  left=5pt,right=5pt,top=5pt,bottom=5pt,
  fonttitle=\bfseries\small,
  title={R2V Prompt Exemplar (Transferring text identity from reference)}
]
\footnotesize
A photorealistic daylight street shot of a boutique storefront. From the first frame, English sign text appears as a single short line of text on a flat signboard above the door, positioned along the top center of the frame, rendered at medium size. It is a single-level text element. The text reads ``Harbor Lane.'' Render it with exactly the glyph shapes, typeface, stroke weight, and color of the lettering in the reference image. The lighting is even and easy to read. The text has high contrast against the background. The camera orbits slowly to the right around the subject, covering about a thirty-degree arc over the clip. The text remains rigidly attached to its real-world surface, undergoing the correct perspective and scale change as the viewpoint shifts, with no drift relative to the surface. There is no occlusion. The text stays fully visible for the entire clip. The text content remains unchanged.
\end{tcolorbox}

\paragraph{I2V.}
I2V uses both synthetic first frames and real images from LSVT~\citep{sun2019icdar} and TextOCR~\citep{9577436}. Annotators transcribe the visible target text, label the visible text properties and imaging conditions, and verify that the source supports the intended temporal behavior. Text scale and position are derived from measured geometry. The temporal prompt is written only after source-side metadata is fixed and specifies scene evolution without restating the full source text or exact font and color identity.

\begin{tcolorbox}[
  enhanced,breakable,
  colback=blue!3,colframe=blue!45!black,
  arc=2pt,boxrule=0.45pt,
  left=5pt,right=5pt,top=5pt,bottom=5pt,
  fonttitle=\bfseries\small,
  title={I2V Prompt Exemplar (Sustaining consistency under dynamics)}
]
\footnotesize
Animate this image as the first frame of the video. The sign text in the upper center of the frame must keep exactly the same characters, glyph shapes, typeface, stroke weight, and color as in the first frame for the entire clip. The scene must continue to evolve as a coherent single shot and must not hold as a still frame. The camera pans smoothly to the right at a constant speed, turning about one third of the frame width over the clip. The text remains rigidly attached to its real-world surface, undergoing the correct perspective and scale change as the viewpoint shifts, with no drift relative to the surface. Around the middle of the clip, a passer-by walks in from the left and briefly covers the right third of the text before clearing it completely. The text content remains unchanged.
\end{tcolorbox}

\paragraph{V2V.}
V2V starts from real source videos. We filter clips by spatial resolution and duration, apply video-only quality screening, segment candidate clips with a shot-boundary detector, and retain text-bearing segments. Expert annotation then records the 12 factors, the source and target text, the localized edit type, and the paired before/after metadata. The exemplar below gives a replacement instruction; deletion, addition, restyling, and translation samples replace its first sentence with the corresponding edit instruction.

\begin{tcolorbox}[
  enhanced,breakable,
  colback=blue!3,colframe=blue!45!black,
  arc=2pt,boxrule=0.45pt,
  left=5pt,right=5pt,top=5pt,bottom=5pt,
  fonttitle=\bfseries\small,
  title={V2V Prompt Exemplar (Editing localized text within video)}
]
\footnotesize
Edit the video to replace the text ``Harbor Road'' on the wall sign with ``Cloud Lane.'' Keep the same font style, size, color, position, perspective, and temporal behavior. Do not alter any other text or visual element in the scene.
\end{tcolorbox}

\begin{figure*}[t]
\centering
\begin{minipage}[t]{0.80\textwidth}
  \centering
  \includegraphics[width=\linewidth]{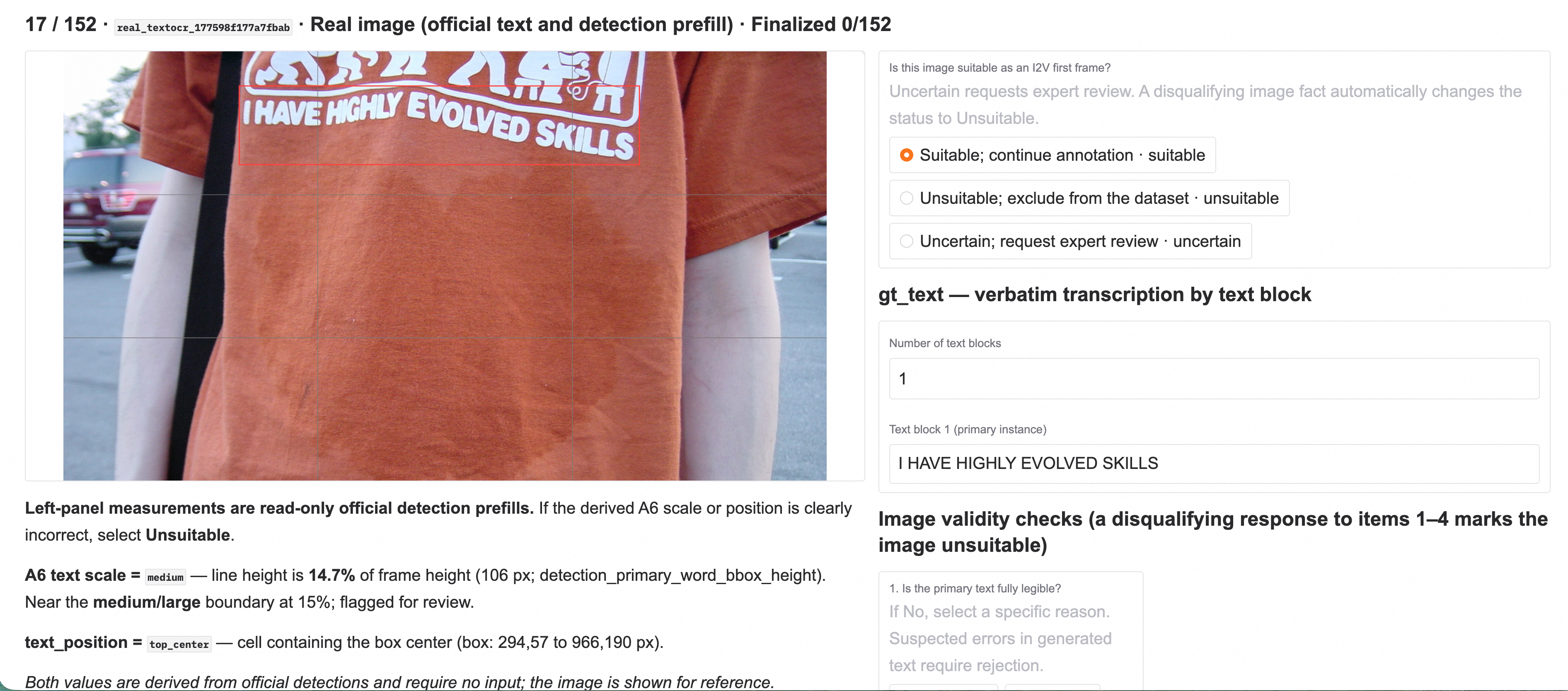}

  \small\textbf{(a)} I2V blind annotation and geometric derivation.
\end{minipage}\hfill
\begin{minipage}[t]{0.80\textwidth}
  \centering
  \includegraphics[width=\linewidth]{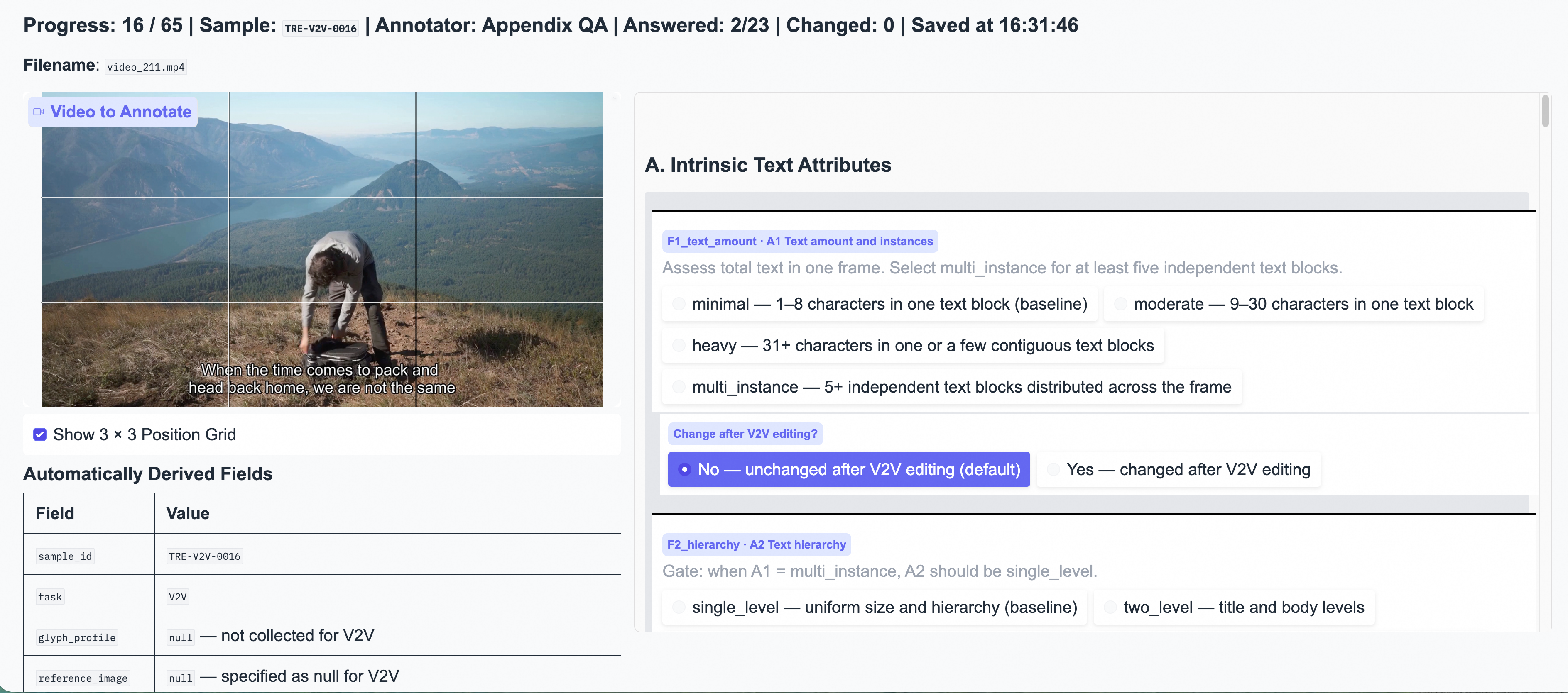}

  \small\textbf{(b)} V2V before/after expert annotation.
\end{minipage}

\caption{Expert annotation interfaces. Sensitive source text and personal information are masked in the visualization.}
\label{fig:app-annotation-interfaces}
\end{figure*}

\paragraph{Expert annotation interfaces.}
Figure~\ref{fig:app-annotation-interfaces} shows the annotation interfaces for I2V and V2V. The I2V interface presents the source image with geometric measurements and annotation fields. The V2V interface presents the source video with paired before/after metadata. Ambiguous samples are escalated to expert review.

\subsection{Conditionally Orthogonal Factor Space}
\label{app:orthogonality}

VidScribe is built on a conditionally orthogonal 12-axis factor space. Here, conditional orthogonality refers to the benchmark construction scheme: each axis is introduced to capture a distinct source of variation, while the instantiation protocol restricts combinations that would be physically undefined, internally inconsistent, or not visually measurable under the benchmark setting. This construction keeps the factor space suitable for controlled diagnostic analysis across tasks.

Not all factor combinations are valid under the benchmark protocol. During construction, we exclude combinations whose difficulty would be dominated by specification artifacts rather than model behavior, as well as combinations that cannot be resolved reliably at the target output format. These decisions are governed by the full construction protocol, which includes clip-level feasibility, rendering scale, metric-facing measurability, font support, and additional task-specific checks. Table~\ref{tab:app-orthogonality-gates} summarizes the principal feasibility constraints used during benchmark construction. Together with the remaining construction checks, these constraints shape the realized benchmark support and make factor-based comparisons more controlled and interpretable.

\begin{table*}[t]
\centering
\caption{Principal feasibility constraints on the joint benchmark support.}
\label{tab:app-orthogonality-gates}
\small\rmfamily
\setlength{\tabcolsep}{5pt}
\renewcommand{\arraystretch}{1.14}
\begin{tabular}{@{}p{0.12\textwidth}p{0.37\textwidth}p{0.43\textwidth}@{}}
\toprule
\textbf{Affected pair} & \textbf{Gate} & \textbf{Reason} \\
\midrule
F1 $\times$ F12
& typing excludes heavy; handwriting\_stroke requires minimal
& The progression must remain visually resolvable within a nominal 5-second clip. \\

F1 $\times$ F2
& minimal permits only single\_level or two\_level
& Minimal text cannot support richer hierarchy patterns. \\

F5 $\times$ F6
& tiny excludes light in specification-driven streams
& At 720p, extremely thin strokes become visually unreliable after encoding. \\

F10 $\times$ F11
& camera\_motion\_screen\_locked permits none or off\_screen\_crop, but not scene-object occlusion
& Scene-object occlusion is incompatible with screen-locked anchoring. \\

F4 $\times$ F5
& In R2V, light is retained only for font families backed by an actual light face
& Unsupported font--weight combinations are excluded from reference construction. \\

\bottomrule
\end{tabular}
\end{table*}

\subsection{Quality Control}
\label{app:filtering}

Quality control operates in three stages: deterministic checks, expert review, and generated-output validation. Deterministic checks enforce conformance of sample metadata records to the schema, task routing, target-string alignment, font and color vocabularies, motion consistency, and feasibility constraints. They also screen prompts for unsafe identifiers, public figures, real brands, protected text, and credential-like number patterns. Expert review verifies scene--domain compatibility, carrier and occluder plausibility, cross-factor consistency, temporal evidence, and prompt naturalness. Generated-output validation verifies that the intended stress condition is visible, that the initial text state is observable, and that shot continuity is preserved.

I2V applies additional source-side gates at the target 720p scale. The main target must be character-level readable and spatially localizable. Samples with unintended extra text, foreground-blocked targets, extremely small secondary text, or unresolved multi-color text are rejected or escalated for expert review. V2V applies additional clip screening on spatial resolution, duration, text visibility, and single-shot continuity. Candidate clips are further filtered with video-only quality criteria and a final manual review.

\section{Metrics Details}
\subsection{Metric Computation Details}
\label{sec:metric-computation}

\textbf{Text Recognition.} A vision-language recognizer supplies block-level transcriptions and axis-aligned bounding boxes for the shared grounding stage. Its JSON output takes the form \texttt{\{"text": <string>, "bbox": [x1,y1,x2,y2]\}}, matching the annotation granularity of VidScribe. Coordinates are predicted on a normalized $[0,1000]$ grid and then rescaled to pixel space. The recognition prompt is given below.

\begin{tcolorbox}[
  enhanced,breakable,
  colback=blue!3,colframe=blue!45!black,
  arc=2pt,boxrule=0.45pt,
  left=5pt,right=5pt,top=5pt,bottom=5pt,
  fonttitle=\bfseries\small,
  title={OCR Prompt}
]
\footnotesize

You are an image OCR detector. Recognize all visible text in the image and return the transcribed text and bounding box for each text block.

\vspace{4pt}
\textbf{[Text Chunking Rules]}

Text across different visual lines must be separated into distinct items.
Within the same visual line, if adjacent text differs noticeably in font size, color, or font weight, it must be split into separate items.
For vertical text, determine grouping by semantics: if consecutive characters form a coherent word, phrase, or sentence, output them as a single item; if they are independent visual text blocks, output them separately.
Continuous text arranged in an arc, circular, or curved layout must be output as a single item.
In cases other than those above, follow the chunking style: continuous text regions sharing the same visual style should be output as a single item; do not split character-by-character or word-by-word.
Ignore purely decorative patterns, separator lines, and non-text textures. Each text block must be output only once.
Text content must be transcribed character-by-character strictly based on the directly visible glyph topology (strokes, contours, and relative positions) in the current image. Do not extrapolate, guess, or hallucinate occluded or blurred characters using semantics, linguistic context, common phrases, or reference texts.
If the topology of a character is visible but cannot be reliably mapped to a known character, you must insert a \texttt{<T>} placeholder at that character's position (e.g., output \texttt{A<T>CD} if perceived as \texttt{A?CD}. Do not complete it semantically, nor skip or delete the position.

\vspace{4pt}
\textbf{[Coordinate Rules]}

The bounding box (bbox) uses a fixed normalized [0, 1000] coordinate system, where the top-left corner of the image is (0, 0) and the bottom-right corner is (1000, 1000), with x directed to the right and y directed downward.
The bbox is the minimum axis-aligned bounding box [x1, y1, x2, y2] enclosing all visible glyphs belonging to the item. All coordinates must be integers satisfying x1 \texttt{<} x2 and y1 \texttt{<} y2.
Items must be sorted following natural reading order: typically top-to-bottom, left-to-right; curved text should follow the flow along the arc.

\vspace{4pt}
\textbf{[Output Format]}

Output valid JSON only. Do not include Markdown blocks, explanations, or extra keys:

items:\texttt{[{"text":"transcribed text","bbox":[x1,y1,x2,y2]},{...}]}
\end{tcolorbox}

\textbf{Common Grounding.} Frame-level text detections are associated into a shared spatio-temporal tracking graph that supplies evidence to all scoring pathways. Tracks are formed by clustering normalized transcriptions and resolving spatial assignment between frames, without box interpolation. A visibility flag identifies detections truncated by the frame boundary, and both content and appearance comparisons are restricted to fully visible instances. Each ground-truth block is matched to at most one track by transcription similarity, with geometric fallback for unreadable glyphs. Voting-based metrics use three requests per item at temperature 0 with a fixed seed of $1234$.

\textbf{A1: Content Accuracy.} For each ground-truth text block $g$, we compare the reference $y_g$ with a consensus transcription $\hat{y}_g$ aggregated over fully visible frames within each observation segment. A gap longer than three frames starts a new segment. After normalization that folds full-width forms and ignores whitespace while preserving case and punctuation, segment-level counts $(\mathrm{TP}_s,\mathrm{FP}_s,\mathrm{FN}_s)$ are weighted by support frames $n_s$:
\begin{equation}
\overline{\mathrm{FP}}_g = \frac{\sum_s n_s\,\mathrm{FP}_s}{\sum_s n_s}, \qquad
\overline{\mathrm{FN}}_g = \frac{\sum_s n_s\,\mathrm{FN}_s}{\sum_s n_s}.
\end{equation}
A block is counted as correct only if both aggregated error counts are zero:
\begin{equation}
c_g = \mathds{1}\!\left[\overline{\mathrm{FP}}_g \le \tau \;\wedge\; \overline{\mathrm{FN}}_g \le \tau\right], \qquad \tau = 0,
\qquad
\mathrm{A1} = \frac{1}{|\mathcal{G}|}\sum_{g \in \mathcal{G}} c_g .
\end{equation}

\textbf{A2: Glyph Correctness.} A structural inspector re-reads each text block from up to $K$ sampled frames using both the full frame and the target quadrilateral. Each defective glyph is marked by an anomaly token. Let $n^{\mathrm{bad}}_{b,k}$ be the number of anomalous glyphs in block $b$ at frame $k$:
\begin{equation}
c_b = \mathds{1}\!\left[\forall k \in \mathcal{K}_b:\; n^{\mathrm{bad}}_{b,k} = 0\right],
\qquad
\mathrm{A2} = \frac{1}{|\mathcal{B}|}\sum_{b \in \mathcal{B}} c_b .
\end{equation}
$\mathcal{B}$ includes both annotated target blocks and unmatched generated text tracks, so defective background text is penalized as well.

\textbf{B1: Content Stability.} For each target, we divide the track into 1-second windows and compute a consensus reading $\hat{y}_w$ for each window. Using the time-weighted modal reading $\bar{y}$ as the anchor, we define
\begin{equation}
d_{\mathrm{char}} = \frac{\sum_w n_w \min\!\left(\mathrm{CER}(\hat{y}_w, \bar{y}),\, 1\right)}{\sum_w n_w},
\end{equation}
where $n_w$ is the number of observed frames in window $w$. The final score is
\begin{equation}
\mathrm{B1} = \frac{1}{|\mathcal{G}|}\sum_{g \in \mathcal{G}} \pi_g \left(1 - d_{\mathrm{char},g}\right) \kappa_g ,
\end{equation}
where $\kappa_g \in \{0,1\}$ is the A1 correctness cap and $\pi_g$ is the effective presence ratio.

\textbf{B2: Appearance Stability.} We score adjacent fully visible frame pairs within each segment, excluding pairs that cross a content switch. For a pair $(a,b)$, optical flow warps the text patch of frame $a$ into the view of frame $b$, and the aligned LPIPS~\citep{lpips} distance gives
\begin{equation}
s_{ab} = 1 - \min\!\left(\mathrm{LPIPS}(\tilde{x}_a, x_b),\, 1\right).
\end{equation}
For target $g$, $r_g$ denotes the valid-pair ratio, and $\rho_g$ caps the appearance score by the observed content continuity:
\begin{equation}
\mathrm{B2} = \frac{1}{|\mathcal{G}|}\sum_{g} \min\!\left(\overline{s}_g,\; \rho_g\right) \cdot r_g .
\end{equation}
We apply this metric only when a target has at least 5 valid pairs and $r_g \ge 0.5$.

\textbf{C1: Color Alignment.} The arbiter scores chromatic agreement with the declared text color on a 4-point rubric. Scores are mapped with $\rho = [0, 0, 0.5, 1]$ and aggregated by the most stringent vote:
\begin{equation}
\mathrm{C1} = \frac{1}{|\mathcal{G}|}\sum_{g} \min_{i \in \{1,2,3\}} \rho\!\left(v_{g,i}\right).
\end{equation}

\textbf{C2: Stroke Weight Agreement.} A weight discriminator classifies target crops into $\{$light, regular, bold$\}$. Majority voting yields $\hat{p}_g$. For $K$ ordered tiers, the normalized agreement is
\begin{equation}
\mathrm{C2}_g = 1 - \frac{\left|p_g - \hat{p}_g\right|}{K - 1} \in \{0,\ 0.5,\ 1\},
\qquad
\mathrm{C2} = \frac{1}{|\mathcal{G}|}\sum_{g} \mathrm{C2}_g .
\end{equation}

\textbf{C3: Typographic Style Matching.} A style discriminator classifies target crops into $\{$monospace, sans, script, serif$\}$, followed by majority vote. A target receives a score of 1 if the voted style matches the declared style and 0 otherwise. The video-level score is the macro average over targets.

\textbf{C4: Carrier Type Attachment.} The arbiter selects the observed carrier of the primary target from a fixed option set: flat two-dimensional surface, three-dimensional surface, freestanding, or handwritten. The score is the correct rate over three blind votes:
\begin{equation}
\mathrm{C4} = \frac{1}{3}\sum_{i=1}^{3} \mathds{1}\!\left[\hat{v}_i = v\right].
\end{equation}

\textbf{C5: Spatial Placement Accuracy.} Each target is scored against its declared position on a $3 \times 3$ partition of the frame. Only the first half of the video is evaluated. The rubric assigns 2 for the correct cell, 1 for an adjacent cell, and 0 for a non-adjacent cell. These are mapped by $\mu = [0, 0.3, 1]$:
\begin{equation}
\mathrm{C5}_g = \frac{1}{3}\sum_{i=1}^{3} \mu\!\left(v_{g,i}\right).
\end{equation}
For multi-level layouts, the video-level score is the macro average over targets.

\textbf{C6: Layout Hierarchy Fidelity.} The arbiter selects the observed hierarchy of the primary target from $\{$single level, two levels, three or more levels, table-like$\}$. The score is the correct rate over three votes.

\textbf{D1: Background Consistency.} We sample up to sixteen frames uniformly over the output video, mask detected text regions, and discard frames whose text area nearly covers the image. A frozen vision-language encoder embeds each remaining frame into an $\ell_2$-normalized feature $f_i$. For T2V, R2V, and I2V, the score is the mean pairwise cosine similarity over the $M$ retained frames:
\begin{equation}
\mathrm{D1} = \frac{2}{M(M-1)}\sum_{1 \le i < j \le M} \cos\!\left(f_i, f_j\right).
\end{equation}

For V2V, we extract normalized background features $f_i^{\mathrm{src}}$ and $f_i^{\mathrm{out}}$ from temporally corresponding source and output frames, with text regions masked in both. Source background preservation is measured over $M$ valid frame pairs:
\begin{equation}
\mathrm{D1}_{\mathrm{V2V}} = \frac{1}{M}\sum_{i=1}^{M}
\cos\!\left(f_i^{\mathrm{src}}, f_i^{\mathrm{out}}\right).
\end{equation}

\textbf{E1: Editing Residue.} This probe detects obsolete source text inside the edit region. For each frame $t$, let $h_t \in \{0,1\}$ denote whether a transcription inside the edit region matches the unmatched old-side content. A residue hit is counted only if it persists for at least two consecutive frames:
\begin{equation}
\mathrm{E1} = \max_t h_t h_{t+1} \in \{0,1\}.
\end{equation}

\textbf{E2: Source Text Preservation.} This probe verifies that undeclared source text remains intact. Each preserved source line is evaluated in one-second windows. Within a window, character alignment gives
\begin{equation}
s_w = \frac{2\,\mathrm{TP}}{2\,\mathrm{TP} + \mathrm{FP} + \mathrm{FN}}.
\end{equation}
For each source line $l$, let $\underline{s}_l$ and $\overline{s}_l$ denote its lower and upper preservation bounds. The reported E2 score averages the lower bounds across source lines:
\begin{equation}
\mathrm{E2} = \frac{1}{|\mathcal{L}|}\sum_{l \in \mathcal{L}} \underline{s}_l,
\qquad
\text{upper} = \frac{1}{|\mathcal{L}|}\sum_{l \in \mathcal{L}} \overline{s}_l .
\end{equation}

Scores are first computed per generated video and then averaged over applicable generated videos for each system.

\subsection{Cross-Backbone Evaluator Analysis}
\label{sec:appendix_vlm_details}

We reran the full evaluation pipeline with Gemini-3.8-Flash~\citep{team2023gemini} in place of Qwen3.7-Plus~\citep{qwen37plus}. The evidence pathways, metric definitions, score mappings, aggregation weights, and expert vision modules were held fixed. Prompt wording and option phrasing were adapted to the instruction-following behavior and rubric sensitivity of each backbone, while preserving the same metric definitions and score mappings.

\begin{figure}[h]
\centering
\includegraphics[width=0.9\linewidth]{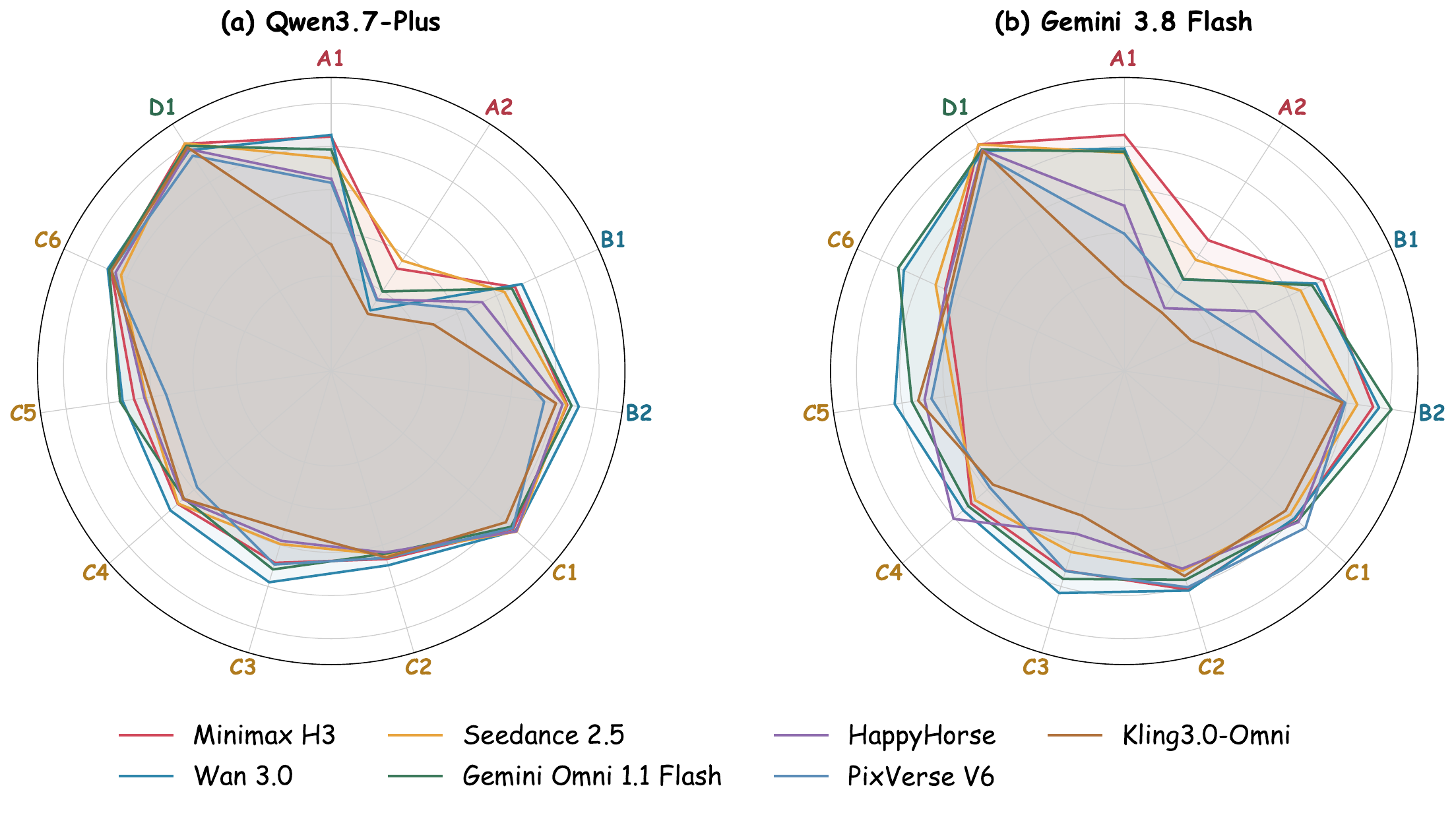}
\caption{Scores of 7 commercial unified video generators across the 11 shared metrics, evaluated with (a) Qwen3.7-Plus and (b) Gemini-3.8-Flash.}
\label{fig:vlm_backbone_radar}
\end{figure}

Figure~\ref{fig:vlm_backbone_radar} compares system profiles across the 11 shared metrics. Gemini-3.8-Flash assigns lower scores on some metrics, but the relative profile of each system remains closely aligned across the two evaluators.

Every system moves by at most one rank under the backbone swap, and the highest- and lowest-ranked systems remain unchanged in overall score. Across these two backbones, the score shift primarily affects absolute calibration rather than comparative system ordering.

\subsection{Human Alignment of Overall System Rankings}
\label{sec:human_overall_alignment}

Human judgments of generated videos integrate task completion with overall visual plausibility. We selected 80 VidScribe samples in approximate proportion to the benchmark task distribution, including 25 T2V, 23 I2V, 14 R2V, and 18 V2V samples. Sample selection prioritized broad coverage of factor axes and difficulty levels within this sample budget. Each of the 7 commercial systems generated one video per selected sample, yielding 560 videos.

8 experts evaluated the videos with model identities hidden. Each video was rated independently by at least 5 experts based on visual quality and task completion, using the output video, prompt, and applicable visual input. Ratings of $1$ to $5$ were mapped to $20$ to $100$. For video $v$, the human score is
\begin{equation}
h_v=\frac{20}{|\mathcal{E}_v|}
\sum_{e\in\mathcal{E}_v}r_{e,v},
\qquad |\mathcal{E}_v|\geq 5,
\end{equation}
where $r_{e,v}\in\{1,2,3,4,5\}$. The task-balanced Human Overall score for model $m$ is
\begin{equation}
H_m=\frac{1}{4}\sum_{t\in\mathcal{T}}
\frac{1}{|\mathcal{V}_{m,t}|}
\sum_{v\in\mathcal{V}_{m,t}}h_v,
\end{equation}
where $\mathcal{T}=\{\mathrm{T2V},\mathrm{I2V},\mathrm{R2V},\mathrm{V2V}\}$. Automated Overall scores were recomputed on the same 560 videos using the 4 group-balanced aggregation and equal task weighting.

\begin{table}[h]
\centering
\caption{Human Overall scores and system rankings on the same 560 generated videos. Overall Rank is determined by the group-balanced VidScribe Overall score.}
\label{tab:human_overall_alignment}
\small
\setlength{\tabcolsep}{10pt}
\begin{tabular}{lccc}
\toprule
System & Human Overall & Human Rank & Overall Rank \\
\midrule
Wan 3.0               & 80.4 & 1 & 2 \\
MiniMax H3            & 80.1 & 2 & 1\\
Seedance 2.5          & 79.5 & 3 & 3 \\
Gemini Omni 1.1 Flash & 79.3 & 4 & 4 \\
HappyHorse            & 77.6 & 5 & 5 \\
PixVerse V6           & 74.3 & 6 & 6 \\
Kling3.0-Omni         & 73.9 & 7 & 7 \\
\bottomrule
\end{tabular}
\end{table}

The VidScribe Overall ranking aligns closely with the human ranking, with only the top two systems exchanging positions. Their Human Overall scores differ by just 0.3\%. This result confirms that the group-balanced Overall score captures holistic human judgments of video quality and task completion.

\subsection{Metric-Level Human Evaluation Protocol}
\label{sec:appendix_human_eval}

\begin{figure}[t]
\centering
\includegraphics[width=\linewidth]{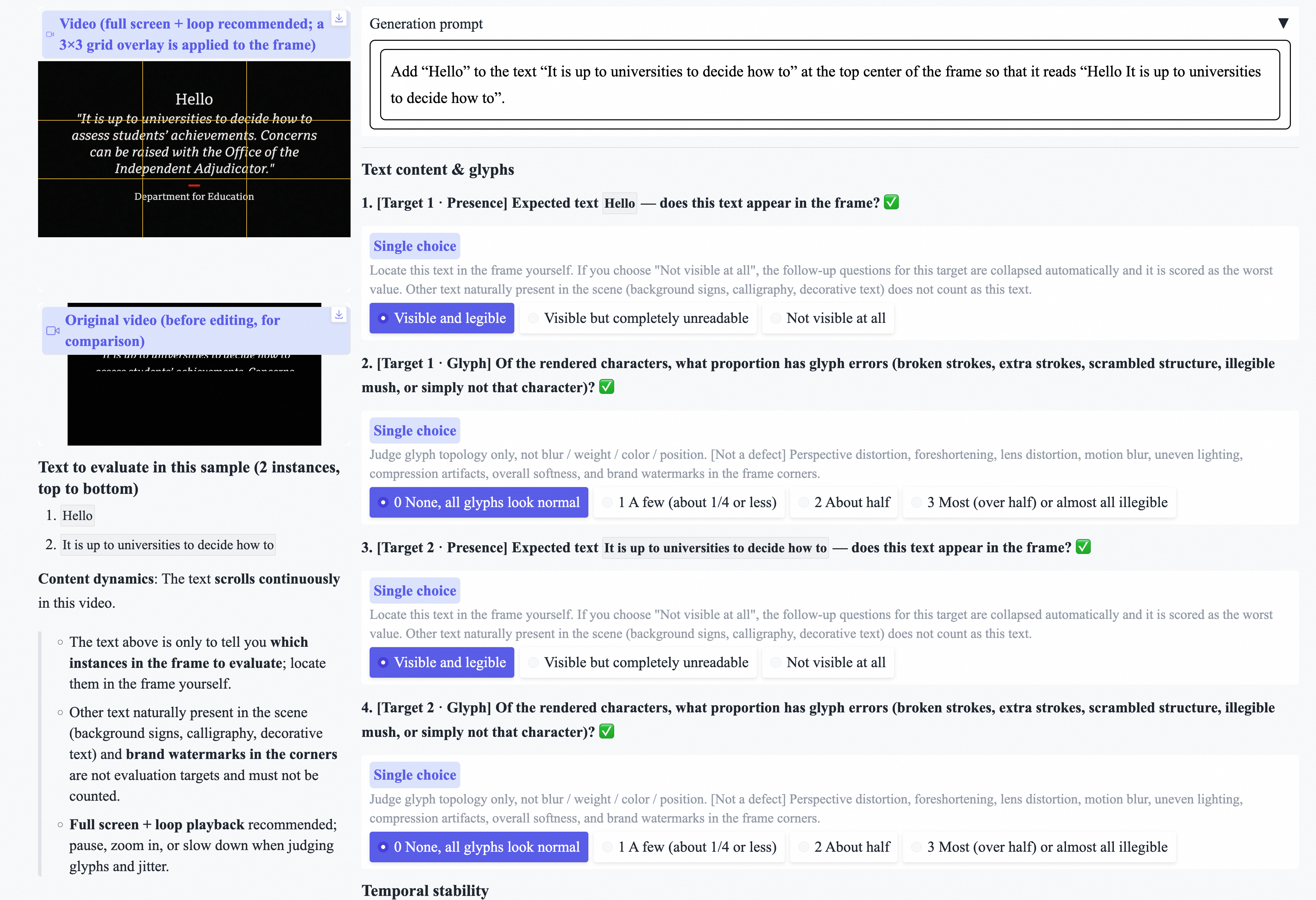}
\caption{Screenshot of the expert questionnaire. Left: the generated video. Right: the survey questions. A $3\times3$ grid is overlaid on the video to assist position judgments.}
\label{fig:human_questionnaire}
\end{figure}

The metric-level human study underlying Table~\ref{tab:human_align} used 950 generated videos sampled uniformly across the 11 systems. 8 experts in typography, graphic design, and video production participated, with each video rated independently by at least 5 experts. The questionnaire provided the expected text and task-specific reference material, including the generation prompt and source video where applicable. Model identities and other raters' responses were hidden. Figure~\ref{fig:human_questionnaire} shows the interface.

Questions correspond one-to-one with benchmark metrics, and their wording and answer options follow the associated evaluation rubrics. The questionnaire uses three response formats:
\begin{itemize}
  \item \textbf{Binary judgments} for pass/fail attributes, such as Editing Residue (E1).
  \item \textbf{Multiple-choice selections} for categorical or relative judgments, such as Typographic Style Matching (C3), Spatial Placement Accuracy (C5), Carrier Type Attachment (C4), and Layout Hierarchy Fidelity (C6).
  \item \textbf{Graded ordinal scales} for text and glyph fidelity, including Content Accuracy (A1) and Glyph Correctness (A2).
\end{itemize}

For each video and metric, majority voting over the 5 independent responses yields the human consensus used in the correlation analysis.

\subsection{Dedicated Discriminators Verification}
\label{sec:appendix_discriminator_verification}

We evaluate the dedicated discriminators for Stroke Weight Agreement (C2) and Typographic Style Matching (C3) on controlled rendered text. Their DINOv2-based design follows prior work on fine-grained font recognition~\citep{parameterdino,dinov2}.

The benchmark renders text from an open-source font collection across styles and weights, then applies curvature perturbation, noise, and background variation. It contains 40{,}000 training images and 10{,}000 test images per task, with balanced class distributions. Both discriminators use a frozen DINOv2 backbone with LoRA~\citep{lora} adapters and are trained separately for style and weight.

The dedicated discriminators achieve 93.88\% Top-1 accuracy on style and 94.67\% on weight, with worst-class accuracies of 88.64\% and 91.52\%, respectively. They outperform both alternatives under both criteria. The difference is especially pronounced for the weakest style class, where Qwen3.7-Plus reaches 14.15\% accuracy. Their advantage in both average and worst-class performance supports their use for C2 and C3 under this evaluation setup.

\begin{figure}[h]
\centering
\begin{subfigure}[t]{0.48\linewidth}
    \centering
    \includegraphics[width=\linewidth]{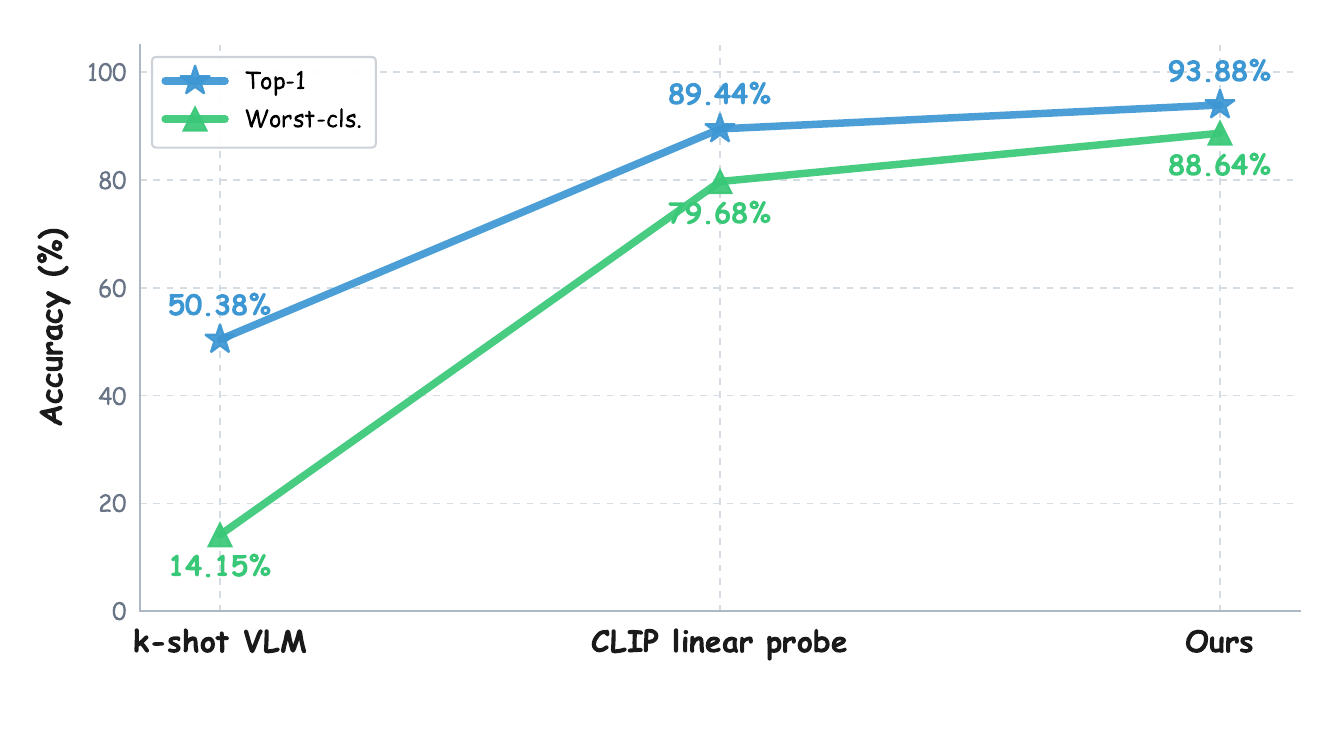}
    \caption{Style ($N_{\mathrm{style}} = 4$ classes).}
    \label{fig:eval_style}
\end{subfigure}
\hfill
\begin{subfigure}[t]{0.48\linewidth}
    \centering
    \includegraphics[width=\linewidth]{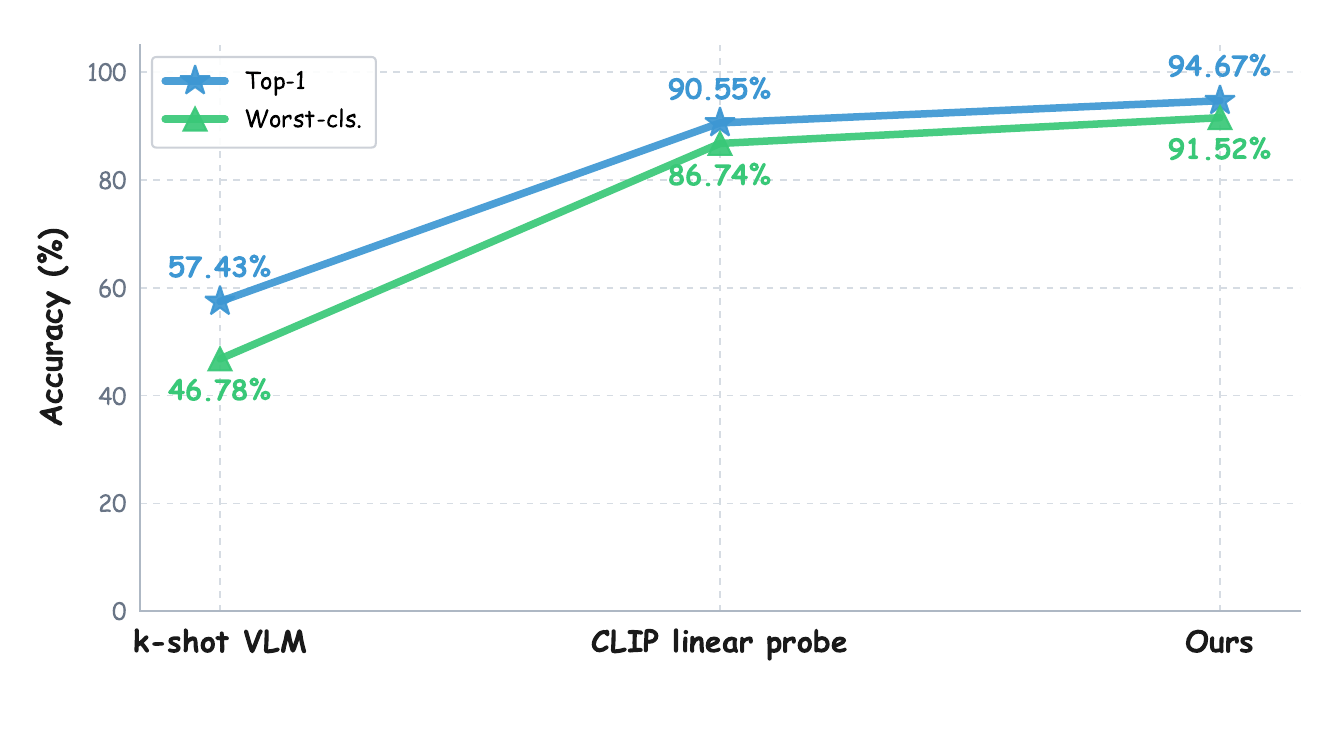}
    \caption{Weight ($N_{\mathrm{weight}} = 3$ classes).}
    \label{fig:eval_weight}
\end{subfigure}
\caption{Comparison of text style and stroke weight classification performance on held-out rendered text. Top-1 denotes overall accuracy, and worst-cls. denotes the lowest per-class accuracy. The VLM baseline uses Qwen3.7-Plus~\citep{qwen37plus}.}
\label{fig:evaluator_selftest}
\end{figure}

\subsection{Metric Sanity Check on Ideal Synthetic Videos}
\label{sec:appendix_metric_sanity}

We build a synthetic validation set of 500 videos at 24 fps using controlled text rendering, background, and motion. The set is organized along VidScribe's factor axes. Target text is rendered from a standard font library and placed on solid-color backgrounds with high text--background contrast, without extra text or occlusion. It covers serif and sans-serif typefaces, regular and bold weights, and variation in text color and position. Each video is generated together with aligned metadata specifying text content, font attributes, color, layout, and transition timing.

We evaluate the set with the full VidScribe pipeline, and Table~\ref{app:tabupper} reports the results. Both text-fidelity metrics reach 100\%, and the temporal-stability metrics reach 100\% and 99.7\%. Across all submetrics, scores range from 90.5\% to 100\%, with an overall score of 97.9\%. These results provide a controlled reference point for the OCR, tracking, and scoring pipeline.

\begin{table}[h]
\centering
\footnotesize
\setlength{\tabcolsep}{2.5pt}
\caption{Metric sanity check on 500 ideal synthetic videos.}
\label{app:tabupper}
\begin{tabular}{l c cc cc cccccc c}
\toprule
& & \multicolumn{2}{c}{\shortstack{\textit{Text}\\\textit{Fidelity}}}
  & \multicolumn{2}{c}{\shortstack{\textit{Temporal}\\\textit{Stability}}}
  & \multicolumn{6}{c}{\shortstack{\textit{Instruction}\\\textit{Compliance}}}
  & \multicolumn{1}{c}{\shortstack{\textit{Non-Text}\\\textit{Preservation}}} \\
\cmidrule(lr){3-4}
\cmidrule(lr){5-6}
\cmidrule(lr){7-12}
\cmidrule(lr){13-13}
Input videos & Overall $\uparrow$
& A1 $\uparrow$ & A2 $\uparrow$
& B1 $\uparrow$ & B2 $\uparrow$
& C1 $\uparrow$ & C2 $\uparrow$ & C3 $\uparrow$
& C4 $\uparrow$ & C5 $\uparrow$ & C6 $\uparrow$
& D1 $\uparrow$ \\
\midrule
Ideal synthetic & 97.9
& 100 & 100
& 100 & 99.7
& 99.6 & 91.6 & 90.5 & 99.5 & 95.2 & 95.7
& 96.2 \\
\bottomrule
\end{tabular}
\end{table}

\section{Additional Experiments and Analyses}
\label{app:additional-analysis}

\subsection{Video-Level A1--A2 Correlation Analysis}
\label{app:a1a2-corr}

We examine the correlation between Content Accuracy (A1) and Glyph Correctness (A2) at the level of individual videos, extending the system-level comparison in Finding~1. Although the two metrics target related aspects of text quality, their different failure criteria and coverage can produce materially different outcomes on individual videos.

We pool all videos from the 7 commercial unified video generators over the full VidScribe benchmark and compute the Spearman rank correlation between each video's A1 and A2 scores. The correlation is weak, with $\rho = 0.246$. In practice, this means that videos with high lexical accuracy often do not achieve equally high glyph correctness. This video-level result sharpens the aggregate pattern in Finding~1 and shows that character recognizability and stroke-level glyph fidelity remain only weakly coupled in current video generation systems.

\subsection{Factor Sensitivity Ranking}
\label{app:axis-damage}

\paragraph{Effect sizes and tests.}
For each factor, the frozen metadata mapping partitions samples into \textit{Hard} ($H$) and \textit{Regular} ($R$) groups. Within each task $t$ we quantify the gap with Cliff's delta, and the factor ranking uses the task-balanced Cliff's delta
\begin{equation}
\delta_t
=\frac{1}{n_Hn_R}\sum_{h\in H}\sum_{r\in R}
\left[\mathbb{I}(h>r)-\mathbb{I}(h<r)\right]
=\frac{2U_H}{n_Hn_R}-1,
\qquad
\delta_{\mathrm{TB}}=\frac{1}{4}\sum_t\delta_t,
\label{eq:cliffs-delta}
\end{equation}
where $\mathbb{I}(\cdot)$ is the indicator function and $U_H$ is the Mann--Whitney statistic~\citep{Mann1947On} for \textit{Hard}, so a negative value indicates that \textit{Hard} samples tend to rank below \textit{Regular} ones, and $\delta_{\mathrm{TB}}$ averages the within-task deltas over T2V, I2V, R2V, and V2V. We attach a $95\%$ task-stratified bootstrap confidence interval to each $\delta_{\mathrm{TB}}$: within every task we resample \textit{Hard} and \textit{Regular} samples with replacement, recompute the per-task deltas and their average, and report the $2.5$/$97.5$ percentiles over 5{,}000 replicates. By contrast, the factor--metric heatmap uses the pooled 803-sample $\delta$, with cell significance tested by the Mann--Whitney $U$ test.

\paragraph{Adjusted effects and multiplicity.}
To assess whether each factor's ranking remains after accounting for the other annotated factors it co-occurs with, we also fit a single regression over all samples,
\begin{equation}
Y_i=\beta_0+\sum_{j=1}^{12}\beta_jF_{ij}+\tau_{\operatorname{task}(i)}+\varepsilon_i,
\label{eq:axis-ols}
\end{equation}
where $F_{ij}$ is the \textit{Hard} indicator for factor $j$ and $\tau$ is a task fixed effect. The model is estimated by HC3-robust OLS. Here, $\beta_j$ captures the score difference associated with factor $j$ after accounting for task and the other 11 factor indicators in the benchmark sample, so a significantly negative $\beta_j$ marks a factor whose degradation signal remains after adjustment for co-occurring conditions. To guard against false positives under many simultaneous comparisons, we report Benjamini--Hochberg~\citep{10.1111/j.2517-6161.1995.tb02031.x} adjusted $p_{\mathrm{FDR}}$ values, which cap the expected false-discovery rate at $q=0.05$ within each family of tests. The correction is applied separately to the 12 factor coefficients (Table~\ref{app:axis-damage-ranking}), the 132 heatmap cells (Figure~\ref{fig:app-axis-metric-heatmap}), and the 8 thresholds in Appendix~\ref{app:cumulative-difficulty}. The threshold curves use 5{,}000 within-bin bootstrap replicates.

\begin{table}[t]
\centering
\caption{Factor sensitivity on 803 VidScribe samples, ranked by task-balanced Cliff's $\delta_{\mathrm{TB}}$ for Overall (\textit{Hard} vs.\ \textit{Regular}; more negative indicates larger degradation). $p_{\mathrm{FDR}}(\beta)$ reports BH-adjusted $p$-values for task- and co-factor-adjusted OLS effects; $^{\ddagger}$ marks $p_{\mathrm{FDR}}(\beta)\geq0.05$.}
\label{app:axis-damage-ranking}
\footnotesize
\setlength{\tabcolsep}{6pt}
\begin{tabular}{@{}c l r r@{}}
\toprule
Rank & Factor & $\delta_{\mathrm{TB}}$ & $p_{\mathrm{FDR}}(\beta)$ \\
\midrule
1 & F12 Content Evolution & $-0.347$ & $<0.001$ \\
2 & F11 Foreground Occlusion & $-0.245$ & $<0.001$ \\
3 & F2 Layout Hierarchy & $-0.208$ & $0.010$ \\
4 & F10 Motion Pattern & $-0.193$ & $<0.001$ \\
5 & F1 Text Amount & $-0.154$ & $<0.001$ \\
6 & F6 Text Scale & $-0.139$ & $0.095^{\ddagger}$ \\
7 & F5 Stroke Weight & $-0.119$ & $<0.001$ \\
8 & F4 Typographic Style & $-0.096$ & $0.042$ \\
9 & F9 Separability & $-0.071$ & $0.135^{\ddagger}$ \\
10 & F8 Illumination & $-0.034$ & $0.511^{\ddagger}$ \\
11 & F3 String Type & $-0.028$ & $0.229^{\ddagger}$ \\
12 & F7 Carrier Type & $-0.028$ & $0.811^{\ddagger}$ \\
\bottomrule
\end{tabular}
\\[3pt]
\end{table}

Table~\ref{app:axis-damage-ranking} lists $\delta_{\mathrm{TB}}$ alongside the BH-adjusted significance of each factor's coefficient $\beta_j$ from Equation~\ref{eq:axis-ols}. With task and the other 11 factor indicators included in the same specification, 7 of the 8 top-ranked factors (F12, F11, F2, F10, F1, F5, and F4) retain significantly negative coefficients. This pattern shows that their ranking is not driven solely by co-occurrence with the other annotated factors in the benchmark; F6 is the lone exception ($p_{\mathrm{FDR}}=0.095$). The 4 tail factors (F9, F8, F3, and F7) do not retain significant coefficients under this specification, consistent with their near-zero $\delta_{\mathrm{TB}}$.

\subsubsection{Factor--Metric Correspondence}
\label{app:axis-metric-correspondence}

\begin{figure*}[t]
\centering
\includegraphics[width=\textwidth]{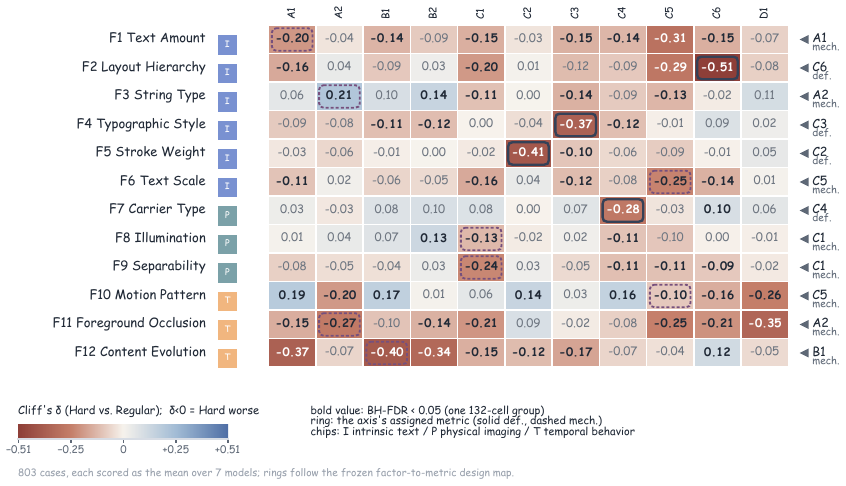}
\caption{Pooled Cliff's $\delta$ (\textit{Hard} vs.\ \textit{Regular}) across VidScribe's 12 factors and 11 shared metrics.}
\label{fig:app-axis-metric-heatmap}
\end{figure*}

Figure~\ref{fig:app-axis-metric-heatmap} uses sample-level scores averaged over the 7 models; every cell includes at least 108 \textit{Hard} and 426 \textit{Regular} samples, the 132 cells form one FDR family, and the color scale is clipped at $\pm0.51$. All 12 assigned cells are significant. The 4 definitional pairs align most strongly with their designated metrics: F2 on C6 Hierarchy Fidelity ($-0.51$), F5 on C2 Weight Agreement ($-0.41$), F4 on C3 Style Matching ($-0.37$), and F7 on C4 Carrier Attachment ($-0.28$). Among the mechanism assignments, the largest mapped effect appears for F12 on B1 Content Stability ($-0.40$). The weakest assigned cell is F10 on C5 Spatial Placement Accuracy ($-0.10$). Overall, 58 cells are significant (49 negative and 9 positive), with the clearest concentration on the temporal-behavior factors.

Several factors also register larger effects on neighboring metrics than on their assigned metrics. F1 Text Amount depresses C5 Spatial Placement Accuracy ($-0.31$) more than its assigned A1 Content Accuracy ($-0.20$), consistent with the added spatial burden imposed by dense text layouts. F11 and F10 have their strongest effects on D1 Background Consistency ($-0.35$ and $-0.26$), a Non-Text Preservation metric not assigned to any text factor. F3 String Type is the only factor whose assigned metric is significantly positive (A2 Glyph Correctness $+0.21$), indicating that its pooled glyph signal differs in direction from the other assigned correspondences once task composition is aggregated. F8 and F9 both map to C1 Color Alignment, so their assigned evidence is partly shared by construction.

\subsection{Benchmark-Aligned Preference Optimization Performance}
\label{app:dpo}

\definecolor{DPOAccent}{RGB}{137,193,142}
\definecolor{DPOTrack}{HTML}{E2E8ED}
\newlength{\DPOBarWidth}
\newlength{\DPOBarUnit}
\providecommand{\DPOWinBar}[1]{%
  \begingroup
  \setlength{\DPOBarWidth}{2.8cm}%
  \setlength{\DPOBarUnit}{0.01\DPOBarWidth}%
  \raisebox{0.1ex}{%
    \makebox[0pt][l]{\color{DPOTrack}\rule{\DPOBarWidth}{1.1ex}}%
    \makebox[0pt][l]{\color{DPOAccent}\rule{#1\DPOBarUnit}{1.1ex}}}%
  \hspace{\DPOBarWidth}\enspace\makebox[2.5em][r]{#1\%}%
  \endgroup}
  
We further evaluate whether VidScribe can function as a practical training signal by applying offline DPO to Wan2.2-TI2V-5B~\citep{wan2025wan} and testing the resulting model on the disjoint VidScribe T2V split.

\paragraph{Preference data.}
We construct 1{,}000 disjoint prompts and sample 8 candidates per prompt from the frozen Wan2.2-TI2V-5B model. VidScribe scores select the highest-Overall candidate as $y^+$ and the lowest as $y^-$, yielding 1{,}000 fixed preference pairs from 8{,}000 videos. VAE latents are cached, and both preference arms share the sampled timestep and noise during training.

\paragraph{Objective and training.}
We define the preference objective by combining a pairwise preference term with a winner-anchor regularizer:
\begin{equation}
\mathcal L
=
-\mathbb E\!\left[\log\operatorname{sigmoid}\!\left(-\beta \omega(\lambda_t)\, m_\theta\right)\right]
+\lambda_{\mathrm{win}}\beta\,\mathbb E\!\left[\omega(\lambda_t)e_\theta(y^+,z_t^+,t\mid x)\right],
\label{eq:dpo-core}
\end{equation}
where $m_\theta$ is the policy--reference margin between the preferred and dispreferred videos, and $e_\theta$ denotes the flow-matching velocity-prediction error. We set $\lambda_{\mathrm{win}}=0.1$ and $\beta=5{,}000$. Training uses transformer LoRA with rank/$\alpha=64/64$ on 16 GPUs with the base model frozen, 121-frame videos at $480\times832$ resolution, AdamW with learning rate $10^{-6}$, a global pair batch size of 16, and cached latents, with CFG disabled.

\paragraph{Automatic evaluation.}
On the 250 held-out T2V prompts with matched sampling, benchmark-aligned DPO improves Overall from 45.92\% to 48.42\% ($+2.50\%$). The gains are concentrated in the benchmark's key structural bottlenecks, with the largest improvements on C6 ($+9.33\%$), A2 ($+5.00\%$), C5 ($+4.62\%$), A1 ($+4.17\%$), C4 ($+4.00\%$), and B1 ($+3.87\%$).

\begin{table*}[t]
\centering
\scriptsize
\setlength{\tabcolsep}{1.7pt}
\renewcommand{\arraystretch}{0.95}
\caption{Wan2.2-TI2V-5B and its benchmark-aligned DPO variant on the VidScribe T2V split. Values and $\Delta$ are percentages and changes.}
\label{tab:dpo-automatic-results}
\begin{tabular}{l c cc cc cccccc c}
\toprule
& & \multicolumn{2}{c}{\shortstack{\textit{Text}\\\textit{Fidelity}}}
  & \multicolumn{2}{c}{\shortstack{\textit{Temporal}\\\textit{Stability}}}
  & \multicolumn{6}{c}{\shortstack{\textit{Instruction}\\\textit{Compliance}}}
  & \multicolumn{1}{c}{\shortstack{\textit{Non-Text}\\\textit{Preservation}}} \\
\cmidrule(lr){3-4} \cmidrule(lr){5-6} \cmidrule(lr){7-12} \cmidrule(lr){13-13}
Model & Overall $\uparrow$ & A1 $\uparrow$ & A2 $\uparrow$ & B1 $\uparrow$ & B2 $\uparrow$
& C1 $\uparrow$ & C2 $\uparrow$ & C3 $\uparrow$ & C4 $\uparrow$ & C5 $\uparrow$ & C6 $\uparrow$
& D1 $\uparrow$ \\
\midrule
Wan2.2-TI2V-5B & 45.92 & 16.83 & 21.90 & 14.50 & 51.23 & \textbf{51.70} & 50.70 & \textbf{27.57} & 44.00 & 41.19 & 54.00 & 86.60 \\
+ benchmark-aligned DPO & \textbf{48.42} & \textbf{21.00} & \textbf{26.90} & \textbf{18.37} & \textbf{51.31} & 49.70 & \textbf{50.85} & 25.90 & \textbf{48.00} & \textbf{45.81} & \textbf{63.33} & \textbf{87.64} \\
$\Delta$ & $+2.50$ & $+4.17$ & $+5.00$ & $+3.87$ & $+0.08$ & $-2.00$ & $+0.15$ & $-1.67$ & $+4.00$ & $+4.62$ & $+9.33$ & $+1.04$ \\
\bottomrule
\end{tabular}
\end{table*}

\begin{figure}[t]
\centering
\includegraphics[width=\linewidth]{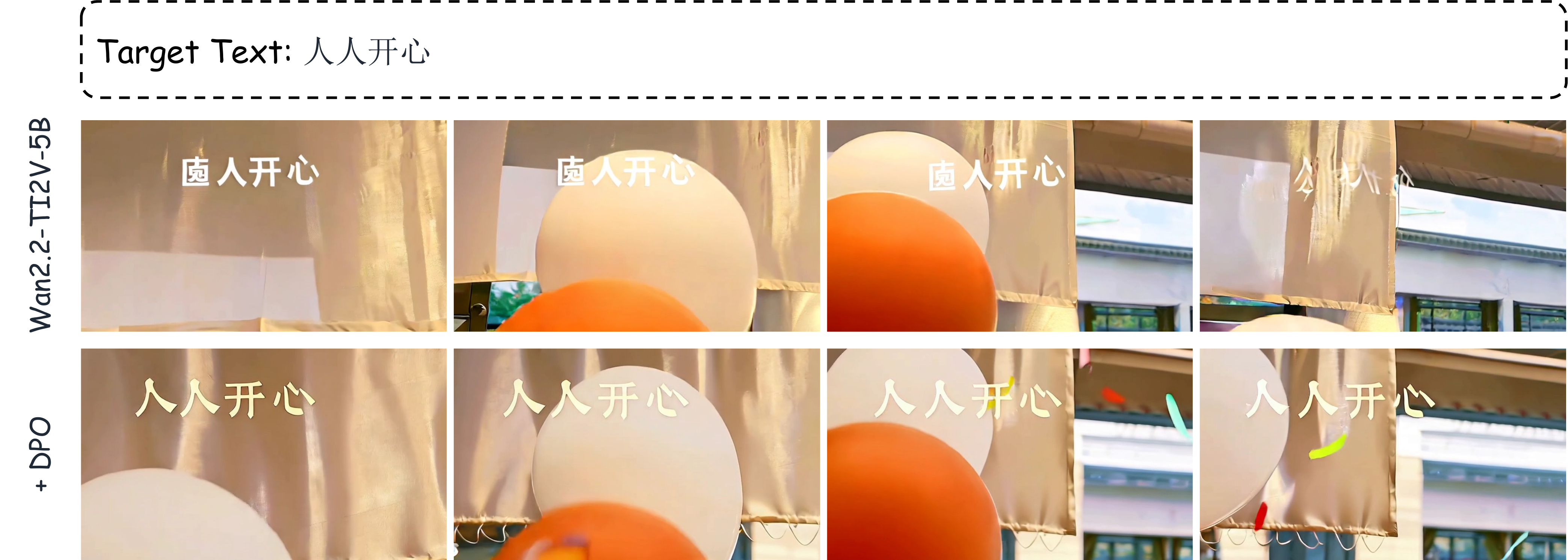}
\caption{Qualitative comparison of glyph rendering by Wan2.2-TI2V-5B (top) and its benchmark-aligned DPO variant (bottom).}
\label{fig:dpo-qualitative}
\end{figure}

\paragraph{User study.}
On the same 250 prompts, benchmark-aligned DPO is preferred to Wan2.2-TI2V-5B on all 4 human-evaluation dimensions. The win rates are 68.8\% for \textit{Text Fidelity}, 54.8\% for \textit{Temporal Stability}, 54.0\% for \textit{Instruction Compliance}, and 51.6\% for \textit{Non-Text Preservation}. Table~\ref{tab:dpo-user-study} summarizes the corresponding vote counts. Together, these results show that VidScribe serves not only as a diagnostic benchmark but also as a practical preference signal for improving video text generation. Qualitative comparisons are presented in Figure~\ref{fig:dpo-qualitative}.

\begin{table}[t]
\centering
\scriptsize
\setlength{\tabcolsep}{4pt}
\renewcommand{\arraystretch}{1.08}
\begin{tabular}{@{} l r l @{}}
\toprule
Dimension & DPO wins & DPO Win Rate \\
\midrule
A: \textit{Text Fidelity} & 172 & \DPOWinBar{68.8} \\
B: \textit{Temporal Stability} & 137 & \DPOWinBar{54.8} \\
C: \textit{Instruction Compliance} & 135 & \DPOWinBar{54.0} \\
D: \textit{Non-Text Preservation} & 129 & \DPOWinBar{51.6} \\
\bottomrule
\end{tabular}
\caption{Blinded user-study results for Wan2.2-TI2V-5B and its benchmark-aligned DPO variant on 250 prompt pairs. Green bars show the DPO win rate in each dimension.}
\label{tab:dpo-user-study}
\end{table}

\subsection{Task-wise Performance}
\label{app:task-wise-model-performance}

Tables~\ref{app:tabt2v}, \ref{app:tabr2v}, \ref{app:tabi2v}, and \ref{app:tabv2v} report the task-wise results on Text-to-video (T2V), Reference-to-video (R2V), Image-to-video (I2V), and Video-to-video (V2V), respectively. The T2V and V2V tables additionally include task-specific open-source systems. These task-wise breakdowns make explicit the capability structure that is compressed by a single aggregate ranking.

\begin{table}[h]
\centering
\footnotesize
\setlength{\tabcolsep}{2.0pt}
\caption{T2V results on VidScribe for commercial unified and task-specific open-source video generators.}
\label{app:tabt2v}
\begin{tabular}{l c cc cc cccccc c}
\toprule
& & \multicolumn{2}{c}{\shortstack{\textit{Text}\\\textit{Fidelity}}}
  & \multicolumn{2}{c}{\shortstack{\textit{Temporal}\\\textit{Stability}}}
  & \multicolumn{6}{c}{\shortstack{\textit{Instruction}\\\textit{Compliance}}}
  & \multicolumn{1}{c}{\shortstack{\textit{Non-Text}\\\textit{Preservation}}} \\
\cmidrule(lr){3-4} \cmidrule(lr){5-6} \cmidrule(lr){7-12} \cmidrule(lr){13-13}
Video generators & Overall $\uparrow$ & A1 $\uparrow$ & A2 $\uparrow$ & B1 $\uparrow$ & B2 $\uparrow$
& C1 $\uparrow$ & C2 $\uparrow$ & C3 $\uparrow$ & C4 $\uparrow$ & C5 $\uparrow$ & C6 $\uparrow$
& D1 $\uparrow$ \\
\midrule
Gemini Omni 1.1 Flash & \best{80.9} & \second{86.0} & 49.7 & \best{83.9} & \best{91.7}
& 85.4 & 70.1 & 74.5 & 59.3 & \best{76.3} & \second{89.7} & \second{92.3} \\
Seedance 2.5 & \second{80.2} & 81.5 & \best{63.4} & 79.4 & 89.7
& 83.2 & 68.9 & 62.0 & 65.7 & 65.3 & 84.1 & \best{92.4} \\
MiniMax H3 & \second{80.2} & 85.0 & \second{56.6} & 82.1 & 88.4
& 79.7 & 70.1 & 68.8 & 63.1 & 69.7 & 87.7 & 91.5 \\
Wan 3.0 & 79.6 & \best{89.6} & 38.7 & \second{83.1} & 88.7
& 81.3 & \best{72.6} & \best{80.6} & \second{68.3} & \second{76.2} & \best{90.5} & 90.2 \\
HappyHorse & 76.6 & 73.3 & 52.3 & 71.1 & \second{90.3}
& \best{87.6} & 66.2 & 56.8 & 59.7 & 70.3 & 86.5 & 91.6 \\
HunyuanVideo-1.5 & 73.7 & 70.4 & 54.3 & 67.7 & 86.3
& 64.2 & 65.5 & 57.4 & 51.3 & 59.7 & 81.7 & 92.1 \\
PixVerse V6 & 71.0 & 61.2 & 44.2 & 59.3 & 84.7
& \second{85.8} & \second{72.3} & \second{77.2} & 57.1 & 52.8 & 87.3 & 87.0 \\
Kling3.0-Omni & 65.9 & 43.2 & 35.5 & 42.4 & 84.9
& 78.6 & 70.3 & 49.3 & \best{69.1} & 64.3 & 88.8 & 90.5 \\
LTX-2.5 & 59.8 & 37.2 & 31.2 & 34.8 & 73.2
& 74.5 & 61.6 & 51.1 & 52.7 & 57.3 & 77.9 & 88.5 \\
\bottomrule
\end{tabular}
\end{table}

\textbf{T2V} shows the tightest competition at the top. Gemini Omni 1.1 Flash achieves the highest overall score at 80.9\%, with Seedance 2.5 and MiniMax H3 close behind at 80.2\%. At the metric level, however, leadership varies by dimension: Gemini Omni 1.1 Flash is strongest on \textit{Temporal Stability} (B1, B2), whereas Wan 3.0 leads Content Accuracy (A1) and several \textit{Instruction Compliance} metrics, including C2, C3, and C6.

\textbf{R2V and I2V} are both led by MiniMax H3, with overall scores of 81.9\% and 84.2\%, respectively, but the advantage is not uniform across metrics. On R2V, Seedance 2.5 and Wan 3.0 follow at 80.9\% and 80.4\%. MiniMax H3 leads A2 and C3, while Wan 3.0 leads A1, B1, C2, and C4. On I2V, Kling3.0-Omni and Seedance 2.5 both reach 83.5\%, within 0.7\% of MiniMax H3. MiniMax H3 leads A1, B1, and D1; Kling3.0-Omni leads B2 and C2; and Wan 3.0 leads C1, C4, and C5.

\textbf{V2V} remains the weakest task overall. Wan 3.0 achieves the highest overall score at 70.3\%, followed by Gemini Omni 1.1 Flash at 69.3\% and MiniMax H3 at 68.7\%. Wan 3.0 leads the \textit{Temporal Stability} metrics and the C3--C4 \textit{Instruction Compliance} metrics, whereas MiniMax H3 leads A1, A2, C1, C2, and D1. Relative to the other three tasks, V2V most clearly separates localized editing ability from the rest of the text-generation pipeline.

The task-wise tables therefore sharpen the aggregated ranking by exposing model strengths and weaknesses at the regime level.

\begin{table}[h]
\centering
\footnotesize
\setlength{\tabcolsep}{2.0pt}
\caption{R2V results on VidScribe for commercial unified video generators.}
\label{app:tabr2v}
\begin{tabular}{l c cc cc cccccc c}
\toprule
& & \multicolumn{2}{c}{\shortstack{\textit{Text}\\\textit{Fidelity}}}
  & \multicolumn{2}{c}{\shortstack{\textit{Temporal}\\\textit{Stability}}}
  & \multicolumn{6}{c}{\shortstack{\textit{Instruction}\\\textit{Compliance}}}
  & \multicolumn{1}{c}{\shortstack{\textit{Non-Text}\\\textit{Preservation}}} \\
\cmidrule(lr){3-4} \cmidrule(lr){5-6} \cmidrule(lr){7-12} \cmidrule(lr){13-13}
Video generators & Overall $\uparrow$ & A1 $\uparrow$ & A2 $\uparrow$ & B1 $\uparrow$ & B2 $\uparrow$
& C1 $\uparrow$ & C2 $\uparrow$ & C3 $\uparrow$ & C4 $\uparrow$ & C5 $\uparrow$ & C6 $\uparrow$
& D1 $\uparrow$ \\
\midrule
MiniMax H3 & \best{81.9} & \second{83.9} & \best{68.5} & 77.6 & \second{89.0}
& 72.2 & \second{84.1} & \best{86.6} & \second{77.7} & 68.6 & \second{74.6} & \second{90.9} \\
Seedance 2.5 & \second{80.9} & 82.9 & \second{65.5} & \second{77.7} & 88.8
& \best{80.7} & 81.7 & 79.9 & 71.7 & 63.7 & 72.4 & \best{91.3} \\
Wan 3.0 & 80.4 & \best{86.0} & 54.7 & \best{81.8} & \second{89.0}
& \second{73.6} & \best{84.8} & \second{85.3} & \best{79.1} & \second{72.8} & 73.6 & 87.7 \\
HappyHorse & 78.3 & 75.6 & 63.7 & 73.3 & \best{90.9}
& 67.9 & 83.4 & 82.4 & 72.7 & 55.3 & 70.3 & 89.7 \\
Gemini Omni 1.1 Flash & 76.3 & 73.3 & 52.4 & 67.9 & 86.0
& 67.6 & 80.5 & 78.8 & 70.3 & \best{73.5} & \best{75.3} & \best{91.3} \\
PixVerse V6 & 75.4 & 71.9 & 60.0 & 66.9 & 87.6
& 72.6 & \second{84.1} & 83.6 & 70.3 & 48.0 & 71.9 & 86.7 \\
Kling3.0-Omni & 71.1 & 61.8 & 51.9 & 55.6 & 81.1
& 70.6 & 83.1 & 81.2 & 59.2 & 62.3 & 73.4 & 87.6 \\
\bottomrule
\end{tabular}
\end{table}

\begin{table}[h]
\centering
\footnotesize
\setlength{\tabcolsep}{2.0pt}
\caption{I2V results on VidScribe for commercial unified video generators.}
\label{app:tabi2v}
\begin{tabular}{l c cc cc cccccc c}
\toprule
& & \multicolumn{2}{c}{\shortstack{\textit{Text}\\\textit{Fidelity}}}
  & \multicolumn{2}{c}{\shortstack{\textit{Temporal}\\\textit{Stability}}}
  & \multicolumn{6}{c}{\shortstack{\textit{Instruction}\\\textit{Compliance}}}
  & \multicolumn{1}{c}{\shortstack{\textit{Non-Text}\\\textit{Preservation}}} \\
\cmidrule(lr){3-4} \cmidrule(lr){5-6} \cmidrule(lr){7-12} \cmidrule(lr){13-13}
Video generators & Overall $\uparrow$ & A1 $\uparrow$ & A2 $\uparrow$ & B1 $\uparrow$ & B2 $\uparrow$
& C1 $\uparrow$ & C2 $\uparrow$ & C3 $\uparrow$ & C4 $\uparrow$ & C5 $\uparrow$ & C6 $\uparrow$
& D1 $\uparrow$ \\
\midrule
MiniMax H3 & \best{84.2} & \best{90.4} & \second{62.5} & \best{85.2} & \second{90.1}
& 92.6 & 76.0 & 77.6 & 83.6 & 78.1 & \second{80.9} & \best{91.3} \\
Kling3.0-Omni & \second{83.5} & 88.0 & 57.6 & \second{84.8} & \best{92.1}
& 92.2 & \best{78.1} & \second{79.5} & 81.5 & \second{78.5} & \best{81.2} & \second{91.0} \\
Seedance 2.5 & \second{83.5} & \second{90.2} & \best{62.7} & 84.4 & 88.5
& 92.0 & 76.0 & 74.8 & 83.3 & 76.6 & 79.9 & 90.6 \\
Wan 3.0 & 81.8 & 87.4 & 54.6 & 82.5 & 88.7
& \best{93.4} & 77.4 & 78.6 & \best{84.9} & \best{78.6} & 80.3 & 88.4 \\
Gemini Omni 1.1 Flash & 81.6 & 87.5 & 56.1 & 81.4 & 86.2
& 91.8 & 74.3 & \best{80.3} & \second{84.4} & 77.5 & 80.6 & 89.3 \\
PixVerse V6 & 80.5 & 88.5 & 55.4 & 79.7 & 82.6
& \second{93.2} & 75.8 & 78.9 & 82.5 & 77.5 & \best{81.2} & 87.3 \\
HappyHorse & 80.4 & 89.2 & 49.2 & 81.1 & 84.9
& 93.1 & \second{77.8} & 77.7 & 84.1 & \second{78.5} & \best{81.2} & 87.3 \\
\bottomrule
\end{tabular}
\end{table}

\begin{table}[h]
\centering
\footnotesize
\setlength{\tabcolsep}{2.0pt}
\caption{V2V results on VidScribe for commercial unified and task-specific open-source video generators.}
\label{app:tabv2v}
\begin{tabular}{l c cc cc cccccc c}
\toprule
& & \multicolumn{2}{c}{\shortstack{\textit{Text}\\\textit{Fidelity}}}
  & \multicolumn{2}{c}{\shortstack{\textit{Temporal}\\\textit{Stability}}}
  & \multicolumn{6}{c}{\shortstack{\textit{Instruction}\\\textit{Compliance}}}
  & \multicolumn{1}{c}{\shortstack{\textit{Non-Text}\\\textit{Preservation}}} \\
\cmidrule(lr){3-4} \cmidrule(lr){5-6} \cmidrule(lr){7-12} \cmidrule(lr){13-13}
Video generators & Overall $\uparrow$ & A1 $\uparrow$ & A2 $\uparrow$ & B1 $\uparrow$ & B2 $\uparrow$
& C1 $\uparrow$ & C2 $\uparrow$ & C3 $\uparrow$ & C4 $\uparrow$ & C5 $\uparrow$ & C6 $\uparrow$
& D1 $\uparrow$ \\
\midrule
Wan 3.0 & \best{70.3} & 62.8 & 34.8 & \best{54.3} & \best{75.3}
& \second{85.8} & \second{69.9} & \best{75.0} & \best{80.9} & 78.7 & \second{87.5} & 88.1 \\
Gemini Omni 1.1 Flash & \second{69.3} & 61.4 & \second{39.6} & \second{52.7} & \second{69.7}
& 78.5 & 69.1 & \second{72.7} & \second{80.7} & 81.0 & 85.9 & 87.5 \\
MiniMax H3 & 68.7 & \best{64.8} & \best{40.4} & 47.0 & 62.0
& \best{86.1} & \best{70.3} & 72.2 & 80.0 & 80.0 & \second{87.5} & \best{88.4} \\
Seedance 2.5 & 64.9 & 49.6 & 38.0 & 40.5 & 61.9
& 79.0 & 67.4 & 71.2 & 80.3 & 78.4 & 84.4 & 87.9 \\
PixVerse V6 & 64.6 & \second{63.8} & 33.9 & 40.0 & 55.1
& 74.4 & 65.2 & 62.9 & 66.7 & \best{84.3} & \best{87.7} & \second{88.3} \\
HappyHorse & 62.3 & 45.3 & 29.9 & 34.0 & 60.4
& 79.8 & 66.5 & 70.4 & 80.2 & 75.0 & 87.3 & 87.7 \\
JoyAI-Video-Edit & 60.1 & 46.9 & 19.6 & 37.6 & 56.1
& 74.3 & 67.6 & 59.8 & 66.9 & \second{82.6} & 87.1 & 87.2 \\
Kling3.0-Omni & 58.9 & 32.1 & 30.9 & 26.5 & 57.5
& 79.5 & 64.7 & 67.2 & 76.6 & 72.3 & 84.4 & 87.9 \\
Kiwi-Edit & 45.0 & 8.4 & 15.1 & 5.5 & 36.0
& 52.0 & 56.8 & 54.0 & 64.4 & 60.3 & 74.5 & 87.0 \\
\bottomrule
\end{tabular}
\end{table}

\subsection{Cumulative Difficulty and Quality Degradation}
\label{app:cumulative-difficulty}

\begin{figure*}[t]
\centering
\includegraphics[width=0.86\textwidth]{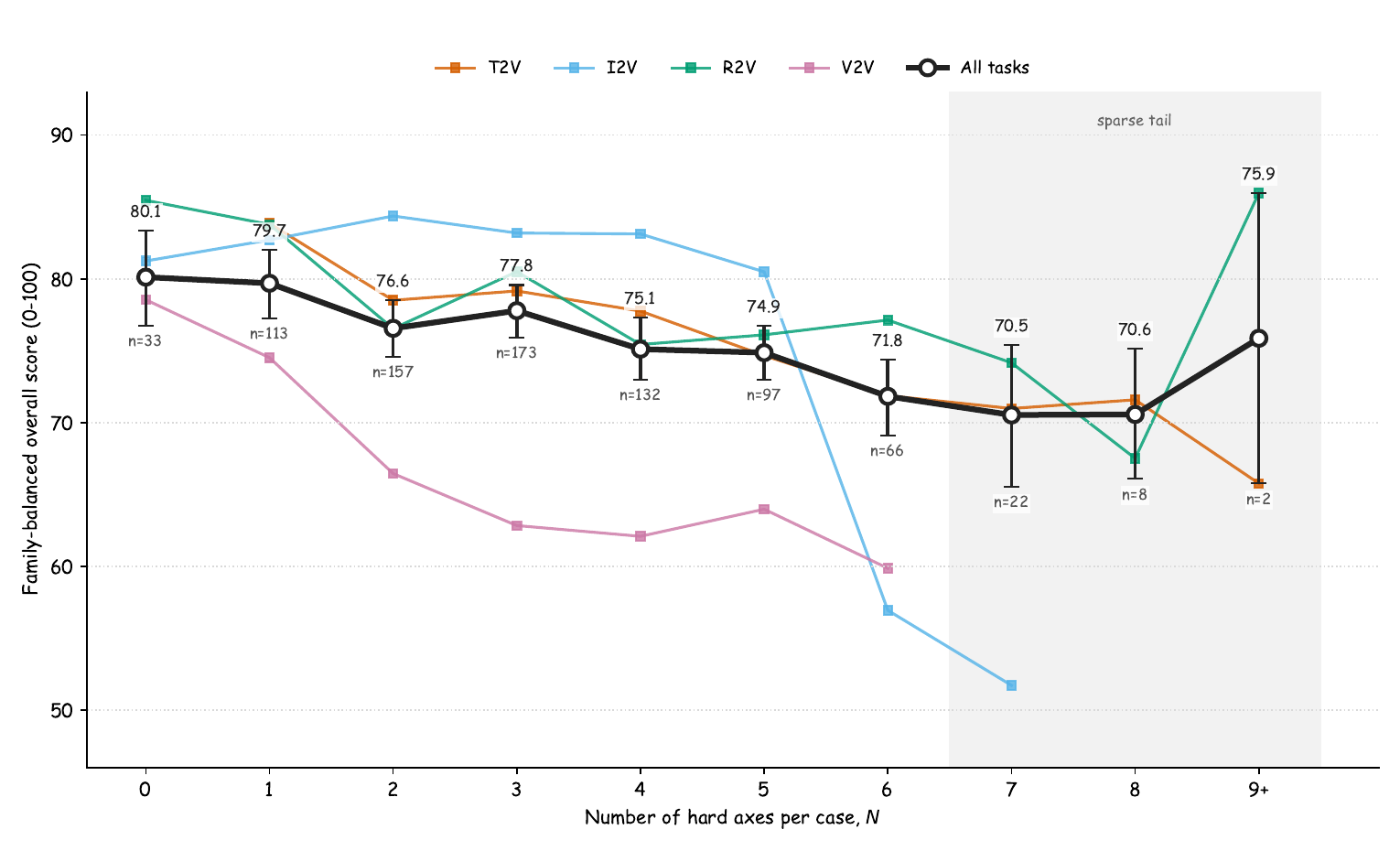}
\caption{Group-balanced Overall scores by the number of \textit{Hard} factors per sample. Black circles show pooled means with 95\% within-bin bootstrap confidence intervals; their labels give the means and sample counts $n$. Colored curves show task-wise means.}
\label{fig:app-cumulative-difficulty}
\end{figure*}

Let $N_i=\sum_{j=1}^{12}F_{ij}$ denote the number of the 12 factors on which sample $i$ takes a \textit{Hard} value. Figure~\ref{fig:app-cumulative-difficulty} shows a clear monotonic trend: as the \textit{Hard}-axis count increases, the group-balanced Overall score declines from $80.1\%$ at $N=0$ to $70.6\%$ at $N=8$. This pattern shows that difficulty accumulates progressively as more \textit{Hard} factors are stacked within the same sample.

The distribution of $N$ is task-dependent, especially in the high-count tail. Mid-range bins contain all 4 tasks, whereas larger counts are concentrated in T2V and R2V. Concretely, $N=6$ is mostly T2V/R2V (with six I2V and two V2V samples), $N=7$ contains T2V/R2V/I2V in $11/9/2$ samples, $N=8$ contains only T2V and R2V in $6/2$ samples, and $N\geq9$ contains two samples in total. Accordingly, the figure reports both the pooled curve and the per-task trajectories, and merges the two $N\in\{9,10\}$ samples into a single $9{+}$ bin.

To test whether the decline contains an additional threshold effect beyond the additive contributions of the individual axes, we augment Equation~\ref{eq:axis-ols} with
\begin{equation}
Y_i=\beta_0+\sum_{j=1}^{12}\beta_jF_{ij}+\gamma\,\mathbb{I}[N_i\geq k]+\tau_{\operatorname{task}(i)}+\varepsilon_i,
\label{eq:threshold-jump}
\end{equation}
and scan $k\in\{2,\ldots,9\}$, correcting the 8 threshold tests as one BH-FDR family. Across all thresholds, no adjusted result is significant. The smallest adjusted value is $p_{\mathrm{FDR}}=0.32$ at $k=9$, where $\gamma$ is positive rather than negative. This agrees with the visual trend in Figure~\ref{fig:app-cumulative-difficulty}: cumulative difficulty is well described by gradual quality degradation, without evidence of a distinct collapse threshold.

\subsection{Failure Mode}
\label{app:qualitative-video-results}

\begin{figure}[htbp]
\centering
\includegraphics[width=0.9\textwidth]{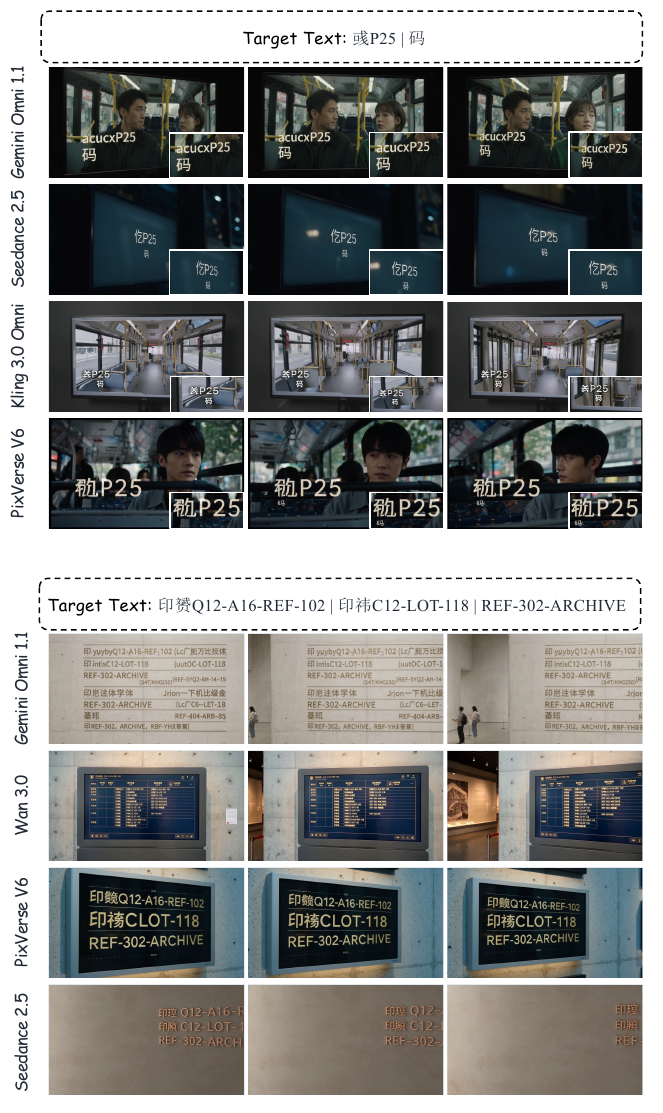}
\caption{T2V failure cases: a rare-glyph target (top) and a multi-line archive sign (bottom).}
\label{fig:app-failure-t2v}
\end{figure}

\begin{figure}[htbp]
\centering
\includegraphics[width=\textwidth]{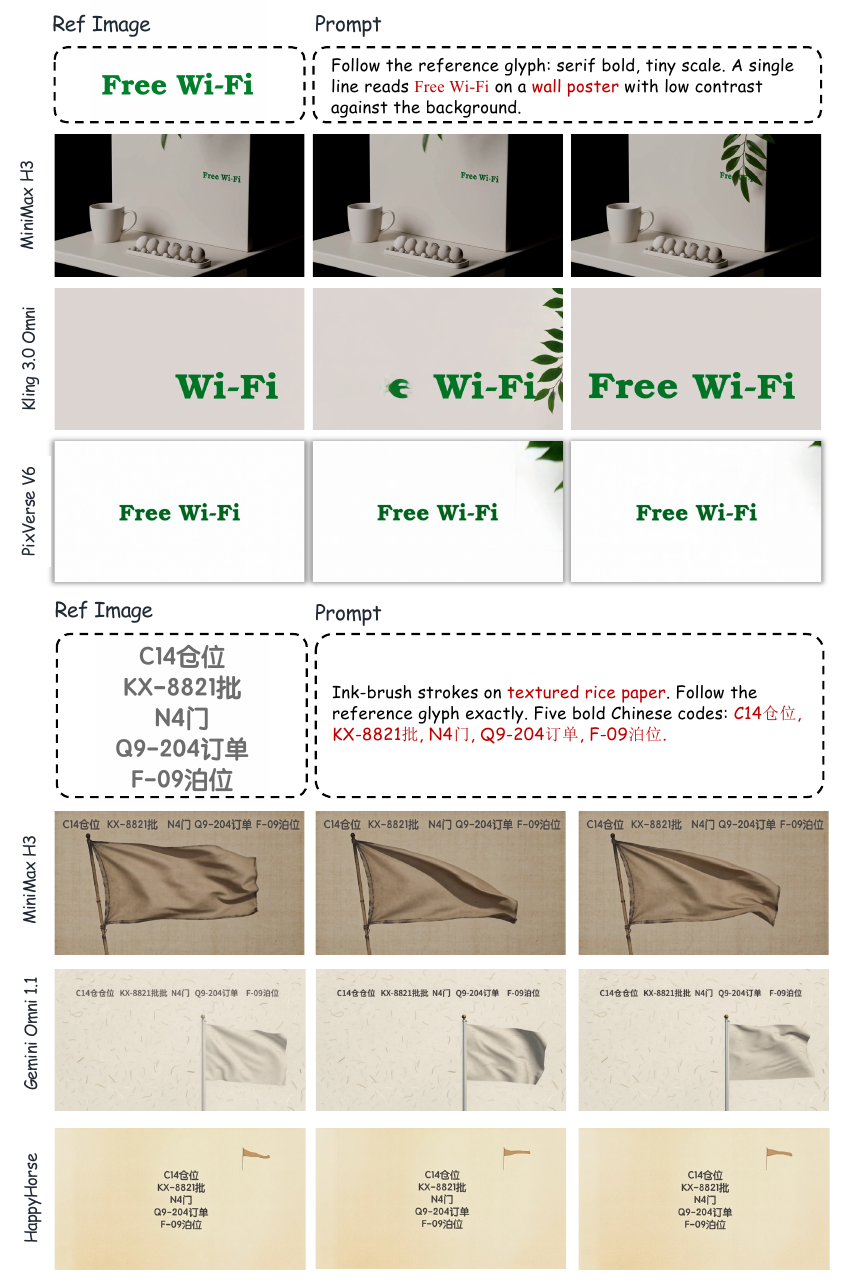}
\caption{R2V failure cases: a low-contrast \texttt{Free Wi-Fi} poster (top, F9) and stacked ink-brush codes (bottom, F5).}
\label{fig:app-failure-r2v}
\end{figure}

\begin{figure}[htbp]
\centering
\includegraphics[width=\textwidth]{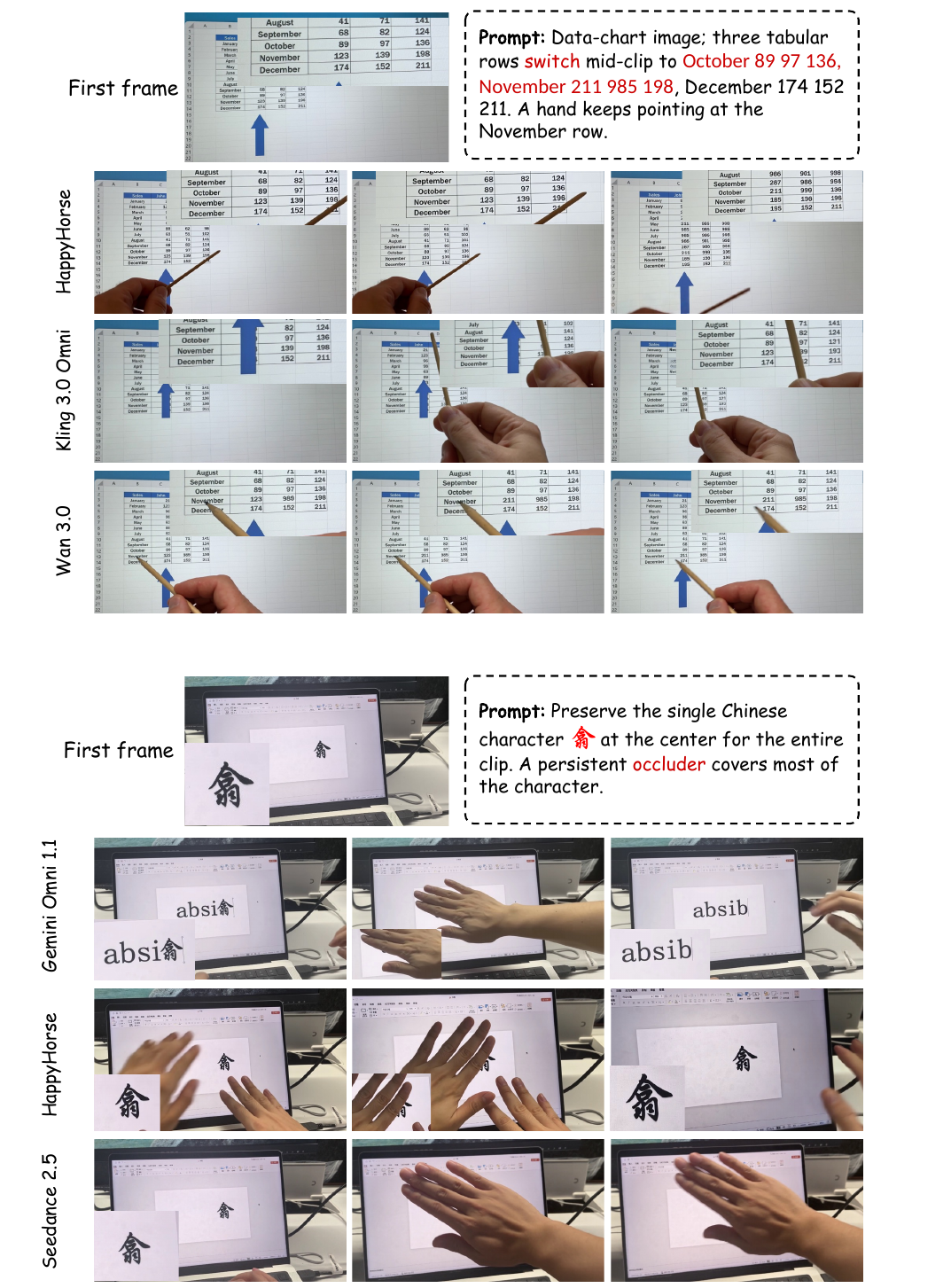}
\caption{I2V failure cases: a mid-clip chart switch (top, F12) and a persistently occluded character (bottom, F11).}
\label{fig:app-failure-i2v}
\end{figure}

\begin{figure}[htbp]
\centering
\includegraphics[width=\textwidth]{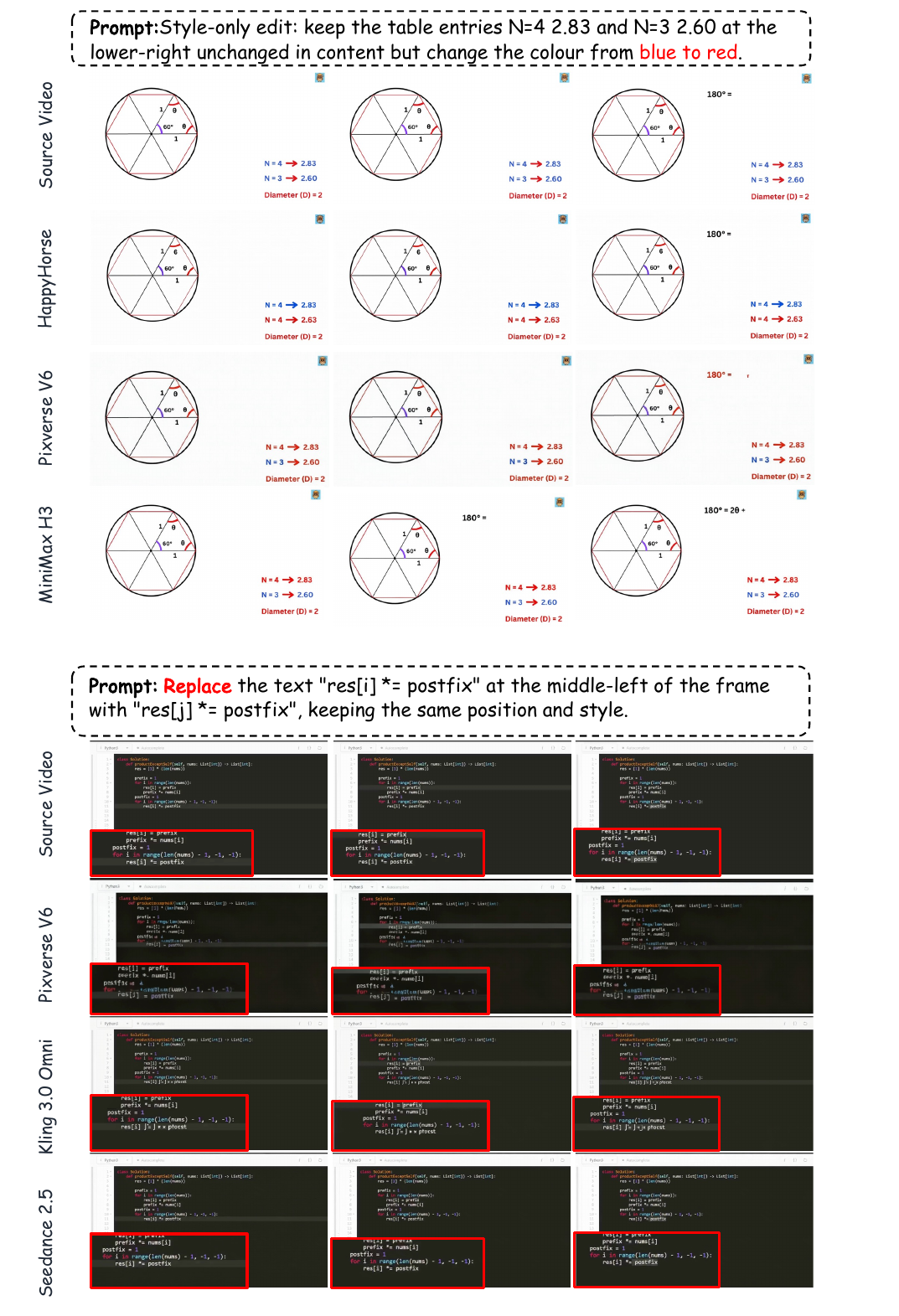}
\caption{V2V failure cases: a style-only recolor edit (top) and a targeted code-line replacement (bottom).}
\label{fig:app-failure-v2v}
\end{figure}

Figures~\ref{fig:app-failure-t2v}--\ref{fig:app-failure-v2v} present one page of failure cases for each generation task, collecting failures that still persist in current systems. Within each case we run several models---including the three overall leaders MiniMax~H3, Wan~3.0, and Seedance~2.5---on the identical prompt or reference so that shared and idiosyncratic failures are directly comparable.

\noindent\textbf{T2V (Figure~\ref{fig:app-failure-t2v}).} For a target that begins with a rare Chinese glyph followed by \texttt{P25}, all 4 models substitute a wrong glyph in the leading position while roughly preserving \texttt{P25}. On a multi-line archive sign, Gemini Omni 1.1 Flash hallucinates a dense block of extra codes, Wan~3.0 collapses the board into an illegible panel, PixVerse~V6 preserves the layout but drops or merges tokens (\texttt{C12-LOT}$\rightarrow$\texttt{CLOT}), and Seedance~2.5 renders only a faint fragment. Unconstrained T2V text therefore fails either through orthographic corruption or through structural over- and under-generation.

\noindent\textbf{R2V (Figure~\ref{fig:app-failure-r2v}).} Following a low-contrast \texttt{Free Wi-Fi} poster (F9), MiniMax~H3 lets the tiny text dissolve into the wall and Kling3.0-Omni drops the \texttt{Free} token for much of the clip, while only PixVerse~V6 keeps the line stable. Given 5 bold ink-brush codes stacked on rice paper (F5), MiniMax~H3 and Gemini Omni 1.1 Flash flatten them into a single thin line and HappyHorse preserves the layout but in a light weight. Models transfer glyph identity from the reference more reliably than they preserve its weight, contrast, or layout.

\noindent\textbf{I2V (Figure~\ref{fig:app-failure-i2v}).} When three chart rows must switch mid-clip (F12) as a hand points at the November row, HappyHorse and Kling3.0-Omni scramble the table and emit wrong numbers, and Wan~3.0 updates only part of the values and only late. Asked to preserve a single Chinese character under a persistent occluder (F11), Gemini Omni 1.1 Flash hallucinates Latin letters beside it, HappyHorse degrades the glyph once the hand lifts, and Seedance~2.5 loses it entirely. A provided first frame does not prevent mid-clip content changes or occlusion from breaking the text.

\noindent\textbf{V2V (Figure~\ref{fig:app-failure-v2v}).} For a style-only edit that should recolor \texttt{N=4 2.83} and \texttt{N=3 2.60} from blue to red without touching content, HappyHorse alters the content (\texttt{N=3 2.60}$\rightarrow$\texttt{N=4 2.63}) while PixVerse~V6 and MiniMax~H3 recolor only part of the entries. For a targeted replacement of \texttt{res[i] *= postfix} with \texttt{res[j] *= postfix} at fixed position and style, PixVerse~V6 and Kling3.0-Omni corrupt the surrounding code and Seedance~2.5 leaves the line unchanged. V2V editing fails through partial application or mis-localization rather than through wholesale regeneration.

Across tasks, these failures cluster into a small set of recurring modes that mirror the benchmark's most damaging axes: wrong or hallucinated glyphs on rare characters, dropped or merged tokens on dense multi-line signs, ignored reference weight and layout (F5), low-contrast text dissolving into its background (F9), scrambled content when the scene changes mid-clip (F12) or is occluded (F11), and partial or mis-localized V2V edits. These failures persist even for the overall leaders---MiniMax~H3 (R2V, V2V), Wan~3.0 (T2V, I2V), and Seedance~2.5 (T2V, I2V, V2V).

\end{document}

%% file: math_commands.tex
\usepackage{amsmath,amsfonts,bm}

\def\eqref#1{equation~\ref{#1}}

\def\1{\bm{1}}

\DeclareMathAlphabet{\mathsfit}{\encodingdefault}{\sfdefault}{m}{sl}
\SetMathAlphabet{\mathsfit}{bold}{\encodingdefault}{\sfdefault}{bx}{n}

\newcommand{\E}{\mathbb{E}}

